%% file: sample-acmtog.tex
\documentclass[acmtog,pbalance,natbib=false]{acmart} 
\AtBeginDocument{%
  }

\makeatletter
\if@ACM@anonymous
  \newcommand{\ifanonymize}[2]{#1}
\else
  \newcommand{\ifanonymize}[2]{#2}
\fi
\makeatother

\RequirePackage[nohyperlinks, printonlyused, withpage, nolist]{acronym}
\RequirePackage{cleveref}
\RequirePackage{threeparttable}
\usepackage{makecell} 
\usepackage{subcaption}
\usepackage{tabularx}
\usepackage{siunitx}
\usepackage{tikz}
\usetikzlibrary{calc}
\usetikzlibrary{positioning}
\usetikzlibrary{calc}
\usetikzlibrary{arrows.meta}

\renewcommand\footnotetextcopyrightpermission[1]{} 
\fancypagestyle{plain}{
  \fancyhf{}
  \fancyfoot[L]{ACM SIGENERGY Energy Informatics Review}
  \fancyfoot[R]{Volume X, Issue X, September 2026}
}

\RequirePackage[
  datamodel=acmdatamodel,
  style=acmnumeric,
]{biblatex}
\begin{document}

\title{Probabilistic Low-Voltage Peak Load Forecasting with Time Series Foundation Models Evaluated on Application-Oriented Metrics}

\author{Benedikt Kaas}
\authornote{Both authors contributed equally to this research.}
\email{b.kaas@netze-bw.de}
\author{Manuel Treutlein}
\authornotemark[1]
\email{manuel.treutlein@partner.kit.edu}
\email{m.treutlein@netze-bw.de}
\orcid{0009-0006-1071-341X}
\author{Hannes Benedikt Gerber}
\affiliation{%
  \institution{Karlsruhe Institute of Technology (KIT)}
  \city{Karlsruhe}
  \country{Germany}
}
\affiliation{%
  \institution{Netze BW GmbH}
  \city{Stuttgart}
  \country{Germany}
}

\author{Oliver Neumann}
\email{o.neumann@netze-bw.de}
\orcid{0000-0003-4438-300X}
\author{Cheewan Phatthanakhuha}
\orcid{0009-0007-8438-7714}
\affiliation{%
  \institution{Netze BW GmbH}
  \city{Stuttgart}
  \country{Germany}
}

\author{Oliver Resch}
\email{oliver.resch@kit.edu}
\orcid{0009-0006-0366-2220}
\author{Ralf Mikut}
\email{ralf.mikut@kit.edu}
\orcid{0000-0001-9100-5496}
\author{Veit Hagenmeyer}
\email{veit.hagenmeyer@kit.edu}
\orcid{0000-0002-3572-9083}
\affiliation{%
  \institution{Karlsruhe Institute of Technology (KIT)}
  \city{Karlsruhe}
  \country{Germany}
}



\begin{abstract}
  Low-voltage load forecasting is an important component in current and future energy systems with a high degree of electrification and decentralized generation. However, current forecasting methods require significant manual effort, often lack uncertainty estimation and proper peak prediction, and they are often not adequately evaluated in terms of grid requirements. In the present study, we provide an extensive evaluation of short-term net load forecasts of $200$ real-world low-voltage feeders with a focus on the rapidly evolving time series foundation models. Our study compares Chronos-Bolt, \mbox{Chronos-2} and TabPFN-TS to six baseline models and demonstrates superior performance, in particular for \mbox{Chronos-2}. An ablation study, in which weather covariates are omitted, shows that time series foundation models adapt to increased uncertainty, despite the importance of weather information. A novel application-oriented metric links the model's forecasting capabilities in peak prediction to the trade-off in grid asset planning and operation between cost reduction and minimizing the risk of failure. 
\end{abstract}






\settopmatter{printacmref=false} 
\maketitle

\thispagestyle{plain} 
\pagestyle{plain} 

\input{acronyms}

\section{Introduction}\label{sec:introduction}

The energy transition in \ac{LV} grids is driven by climate goals and the economic advantages of electrification and private \ac{PV} generation. However, the full-scale integration of electric vehicles, heat pumps, air conditioners, batteries, and \acp{PV} presents new challenges to \acp{DSO} in planning and operation~\cite{cakmakUsingOpenData2022}. In this context, load forecasts of \ac{LV} feeders based on \ac{AI} could enable \acp{DSO} to efficiently identify critical future overload situations. For example, in the German context, this could support congestion management under §14a EnWG while reducing computational overhead. Furthermore, setting dynamic grid fee prices or choosing flexibility options in the next hours and days can profit from forecasts predicting the grid capacity in the near future. A short-term forecasting horizon of four days, as employed in the present study, is thus a reasonable and practical choice. In time series forecasting, a paradigm shift can be observed towards \acp{TSFM}, inspired by the advances in \acp{LLM}. They also offer a promising approach to \ac{LV} load forecasting by enabling zero-shot or few-shot inference, thereby minimizing the need for training pipelines. However, concerns about their general robustness, combined with insufficient evaluation in this area, justify careful and thorough assessment, particularly considering critical infrastructure applications.

To the best of our knowledge, short-term time series forecasting of \ac{LV} feeders, including recent \acp{TSFM} is missing in the literature. In general, load forecasting at the \ac{LV} feeder level is rarely studied compared to \ac{LV} forecasting of smart meter values of households and buildings~\cite{habenShortTermLoad2019}. Furthermore, including weather information as covariates and providing probabilistic, peak-focused \ac{LV} load forecasts is an important research direction~\cite{habenReviewLowVoltage2021}. In addition, \acp{DSO} have special requirements for forecasts regarding peak performance, which are still insufficiently addressed in the literature.

The key contributions in the present paper comprise: 
\begin{enumerate}
    \item We provide an extensive evaluation of several \acp{TSFM} (Chro\-nos-Bolt, \mbox{Chronos-2}, TabPFN-TS) on \ac{LV} load forecasting. The used \acp{TSFM} are selected from an extensive list based on several criteria relevant for \ac{LV} load forecasting. Our study uses a recently published open dataset \ifanonymize{and}{(FeederBW) and} compares the results with fully trained models (tree-based, neural networks) and a naive baseline.
    \item We investigate the role of covariates for load forecasting by providing a short ablation study regarding weather covariates. 
    \item We propose a novel application-oriented metric for evaluating the load forecasting of \ac{LV} feeders in an application-oriented manner. Our metric is inspired by the characteristic curve of the fuse and results in a percentage \ac{KPI} which can be used in management and steering boards of \acp{DSO}.
\end{enumerate}

Overall, this work provides an extensive evaluation of \acp{TSFM} in the context of load forecasting of \ac{LV} feeders, which helps \acp{DSO} and researchers in \ac{LV} load forecasting to better assess the capabilities of this recent model class. The probabilistic nature of the forecasts is implemented as quantile forecasts. The results are based on real-world data, which was recently published in \cite{treutleinRealworldEnergyData2026} and after the release of the examined foundation models. Thereby, we rule out information leaks from foundation models that may have used the dataset during training.  Furthermore, we show that forecasts of \acp{TSFM} are significantly improved by using covariates, which can be well illustrated due to the dependence of \ac{PV} generation on solar irradiance. Our proposed metric helps assess whether the \ac{LV} load forecasts are suitable for use at \acp{DSO} by mapping the technical requirements of exceeding the fuse size into classical data science metrics. We are convinced that in an era of foundation models and highly automated \ac{ML}~\cite{meisenbacherAutomationLevelTaxonomy2025}, application-oriented evaluation plays a decisive role.

The remainder of the present paper is structured as follows. First, we present related literature in \Cref{sec:related_work} which is extended in the appendix in \Cref{sec:time_series_foundation_models} with an overview about different \acp{TSFM} and their components and capabilities. The methodology in \Cref{sec:methodology} is divided into the three key contributions and followed by the experimental setup in \Cref{sec:experimental_setup} about the used dataset, models, and procedure. Analogous to the methodology, we orient the results in \Cref{sec:results} and the discussion in \Cref{sec:discussion} towards the key contributions of the present paper and conclude in \Cref{sec:conclusion}.

\section{Related Work}\label{sec:related_work}

The review in \cite{habenReviewLowVoltage2021} provides a comprehensive overview of \ac{LV} load forecasting until the year 2021. The models that can be used are diverse and comprise statistical methods such as ARIMA and \ac{ML} approaches such as regression trees, kernel-based methods, \acp{MLP} and \acp{DNN}. The authors encourage ongoing research for probabilistic methods, peak forecasting, and leveraging weather information. Indeed, recent research in \ac{LV} load forecasting is increasingly focusing on probabilistic forecasting to address uncertainties inherently present in load and generation of \ac{LV} feeders. For example, uncertainty is originating from \ac{PV} generation~\cite{werlingLinerestrictedDispatchableFeeders2022} and human behavior, such as the time of electric vehicle charging~\cite{amara-oualiQuantifyingUncertaintyElectric2025}. Probabilistic approaches can be based on simple \acp{MLP}~\cite{faustineEfficiencySimplicityMLPBased2025}, Normalizing Flows~\cite{arpogausShortTermDensityForecasting2023, heidrichUsingConditionalInvertible2024} or ensembling with \acp{DNN}~\cite{caoHybridEnsembleDeep2020}. While the present study represents uncertainty using quantiles, probabilistic forecasts can also be expressed through prediction intervals, fully estimated distributions, or ensembles~\cite{habenReviewLowVoltage2021}. In general, quantile and interval forecasts are less computationally expensive and easier to interpret and communicate than fully estimated predictive distributions~\cite{habenReviewLowVoltage2021}.

Probabilistic forecasting also provides better information regarding critical peak values. Compared to other time series forecasting tasks, peak estimation is crucial in \ac{LV} grids because \acp{DSO} plan and operate the grids according to the peak loads. Daily peak prediction is addressed in \cite{gilbertProbabilisticLoadForecasting2023} with semi-parametric \ac{GAMLSS} for smart meter values and the central task in the BigDEAL challenge regarding magnitude, timing, and shape for \ac{MV}/\ac{LV} transformer load~\cite{shuklaBigDEALChallenge20222024}. The design of loss functions and evaluation metrics better reflecting peak performance is a viable research topic, for example, to reduce the double-penalty effect~\cite{habenNewErrorMeasure2014}. Furthermore, novel metrics have been proposed to better capture application-specific forecasting requirements. A notable example is the ramp rate metric introduced in \cite{nouriRampRateMetric2024}, which evaluates ramp events in solar irradiance forecasts using accuracy, precision, recall, and F1 score. Regarding covariates, the study in \cite{habenShortTermLoad2019} about short-term load forecasting of 100 \ac{LV} feeders concludes that temperature has little effect on forecast accuracy or is even worsening. This could potentially be due to the seasonal correlation being more decisive than the temperature correlation. However, weather covariates are still crucial in \ac{LV} forecasting, in particular in the face of high \ac{PV} penetration~\cite{moreno-munozShortTermForecasting2008}. 

Recently, \acp{TSFM} show promising results in time series forecasting while at the same time being much easier to deploy and use since model training becomes largely obsolete. In \Cref{sec:time_series_foundation_models} in the appendix, we provide more background information about different \acp{TSFM} including their architectures and characteristics. A preprint for short-term household electricity load forecasting showed that \acp{TSFM} outperformed other benchmark models even though the latest models were not yet available at the time of the study~\cite{meyerBenchmarkingTimeSeries2025}. Live benchmarks are helpful tools to observe the currently best performing \acp{TSFM}~\cite{meyerTSArenaTechnicalReport2025}. Hence, the obvious question arises whether \ac{LV} load forecasting also profits from this new model class. A parallel work evaluates point load forecasting on three datasets representing different grid aggregation levels, including recent \acp{TSFM}~\cite{hertelDACH+2026}. However, we focus in the present paper on probabilistic load forecasting and evaluate with respect to peaks through a novel application-oriented metric.

\section{Methodology}\label{sec:methodology}

In the present study, we follow a typical \ac{ML} pipeline comprising data preparation, model training and inference, and model evaluation. Although the focus lies on the zero-shot performance of \ac{TSFM}, model training is still required from scratch for the fully trained benchmark models. A standard train-test split is applied for model training and evaluation, which requires a dataset with a sufficient amount of \ac{LV} feeder load measurements. We rule out the possibility that the \acp{TSFM} have already been trained on the data. The following subsections detail the methods used for each key contribution stated in \Cref{sec:introduction}.

\subsection{Model Selection}

To provide an extensive evaluation of the performance of \acp{TSFM}, several benchmark models with different capabilities are required, including a naive baseline model, a tree-based model, and a neural network approach, all of which must provide quantile estimates. The \acp{TSFM} are selected from \Cref{tab:overview_timeseries_foundation_models} in the appendix based on several criteria relevant in \ac{LV} load forecasting, as depicted in \Cref{tab:selection_timeseries_foundation_models}. These criteria comprise that the model weights are publicly available, the model provides probabilistic forecasts, and it can process past and future covariates. Furthermore, it is specified whether fine-tuning code is available, even though fine-tuning is not used in this work. Six out of 26 \acp{TSFM} support probabilistic forecasts and covariates, at least with a covariate regressor, and provide open weights. After several pre-studies with Moirai and Moirai-MoE, these models were excluded due to their limited covariate support capabilities. Regarding the Chronos model family, the most recent models, \mbox{Chronos-2} and Chronos-Bolt, are chosen as they adequately represent the model family.

\input{tables/table_selection_timeseries_foundation_models}

\subsection{Covariates Investigation}

The dataset for our experiment should incorporate significant \ac{PV} generation, which depends heavily on solar radiation and serves as a good prerequisite for an ablation study. In an ablation study, weather covariates, including solar radiation, are provided in one case and omitted in the other. This approach allows us to compare the metric results of the different model predictions. We can expect that a model effectively using the solar radiation covariate can reduce the uncertainty in its forecast, while a prediction without the information about solar radiation should be forced to provide a larger range of uncertainty. While the \acp{TSFM} are not fine-tuned in the present study, a covariate regressor is fitted for Chronos-Bolt to incorporate covariates as suggested in~\cite{abdulfatiransariFastAccurateZeroshot2024}.

\subsection{Calculation of Application-Oriented Metric}\label{subsec:calculation_of_application_oriented_metric}

\begin{figure*}[htb]
    \begin{subfigure}[t]{0.49\linewidth}
    \centering
        \includegraphics[width=\linewidth]{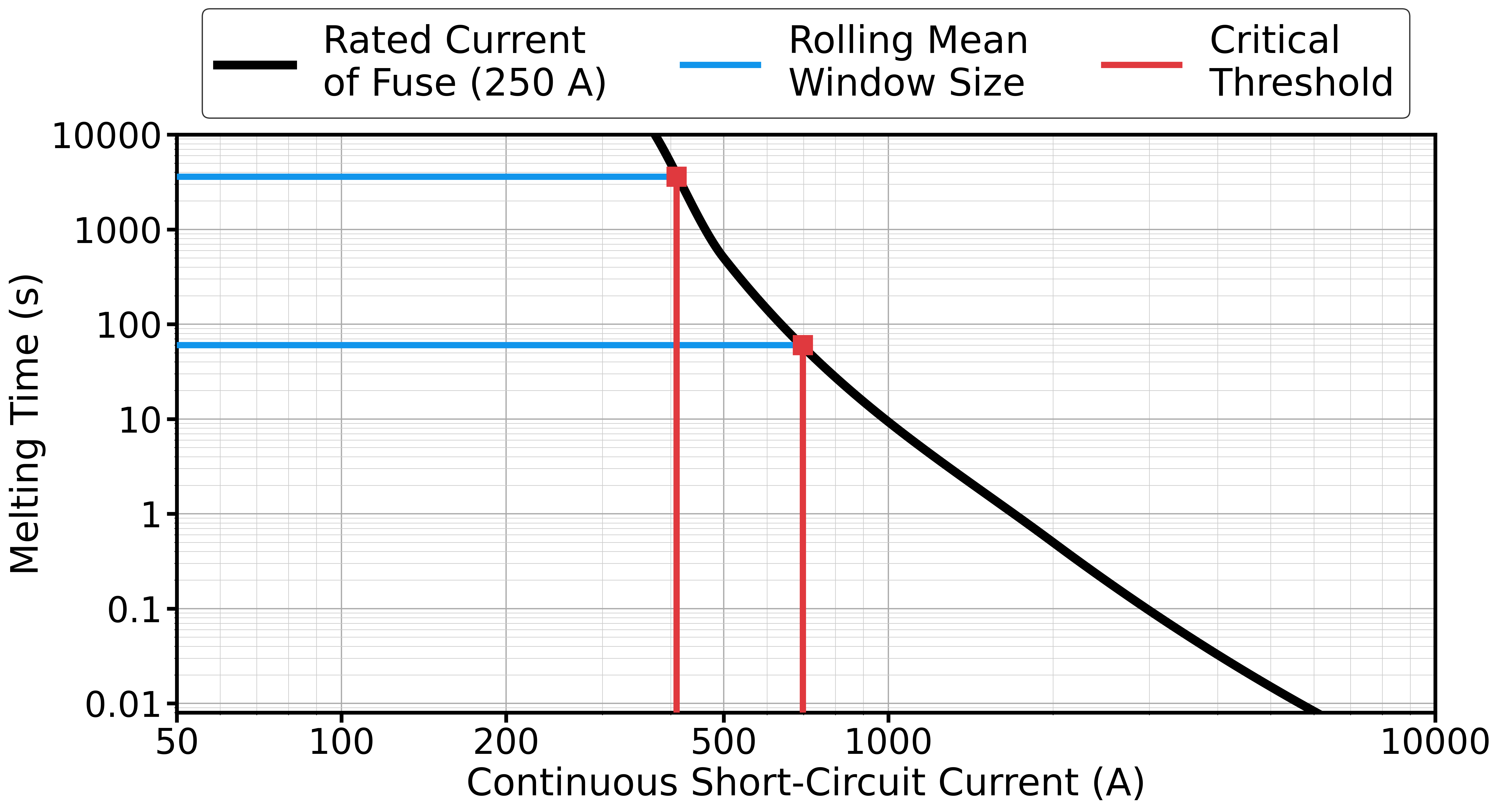}
        \caption{\textbf{Derive parameters} from the time-current curve (window size and threshold).}
        \label{subfig:metric_visualization_application_oriented_1}
    \end{subfigure} \hfill
    \begin{subfigure}[t]{0.49\linewidth}
    \centering
        \includegraphics[width=\linewidth]{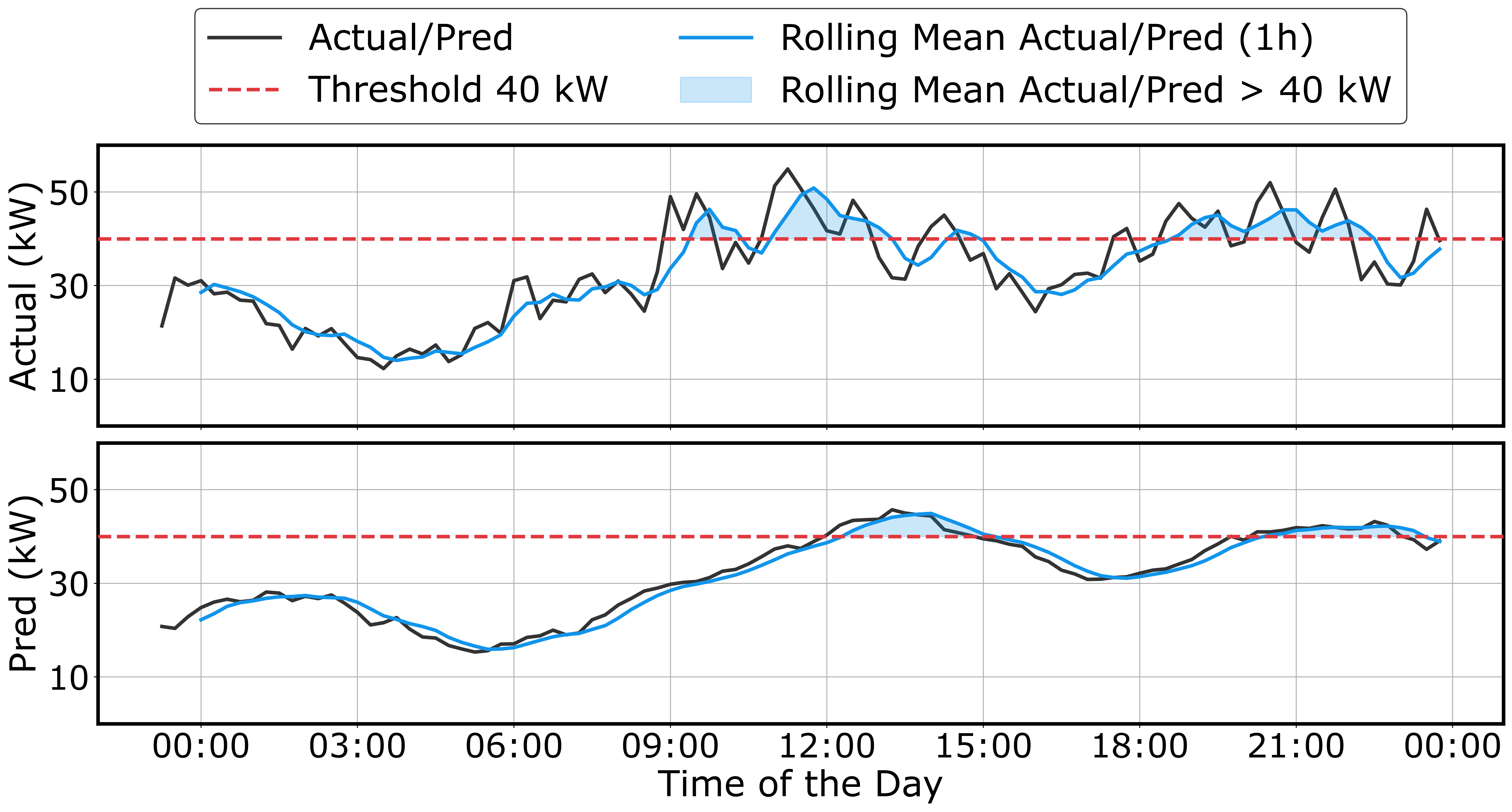}
        \caption{\textbf{Detect critical points} based on rolling mean and threshold.}
        \label{subfig:metric_visualization_application_oriented_2}
    \end{subfigure}
    \begin{subfigure}[b]{0.49\linewidth}
        \centering
        \includegraphics[width=\linewidth]{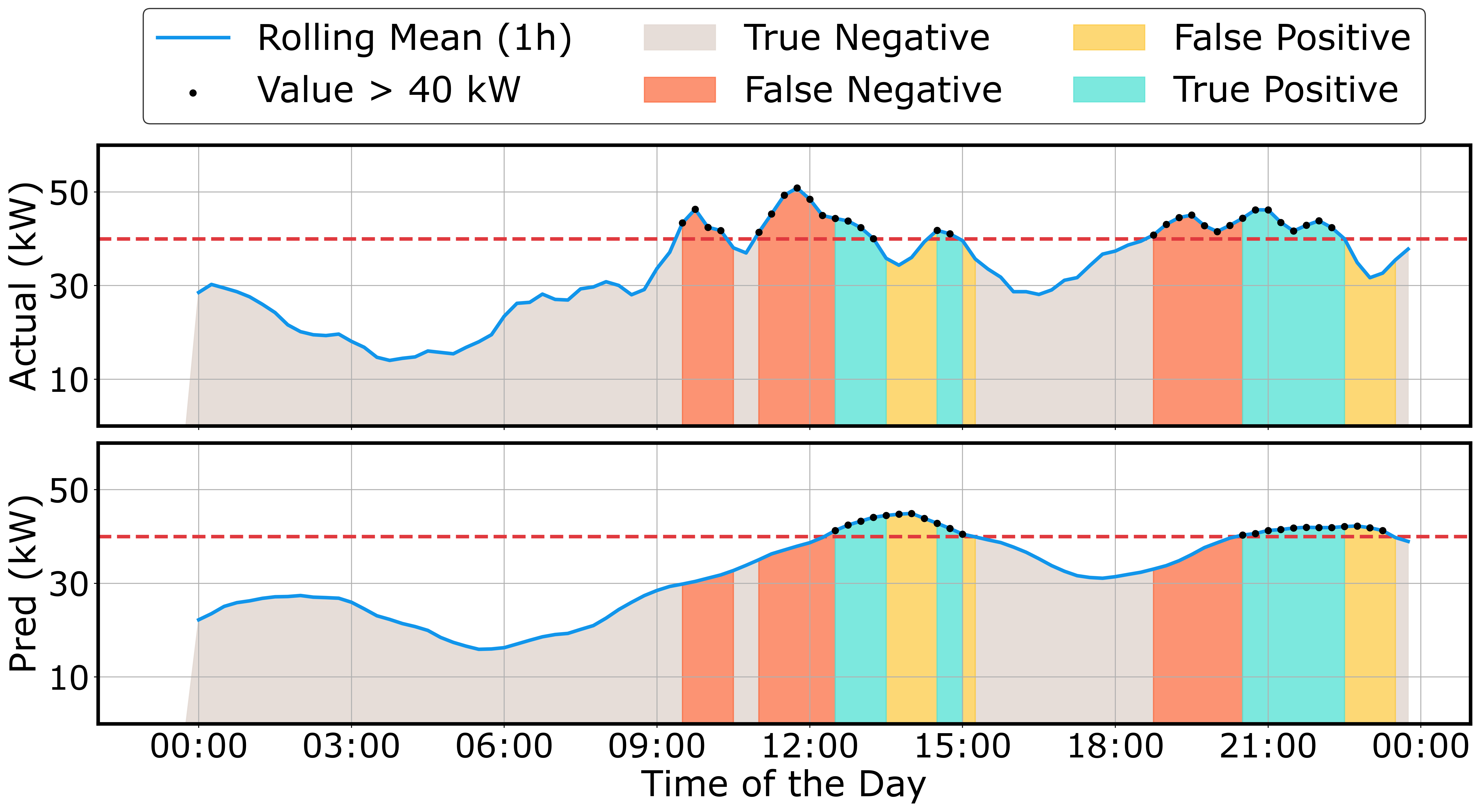}
        \caption{\textbf{Compare critical points} of actual with prediction.}
        \label{subfig:metric_visualization_application_oriented_3}
    \end{subfigure} \hfill
    \begin{subfigure}[b]{0.49\linewidth}
        \centering
        \includegraphics[width=\linewidth]{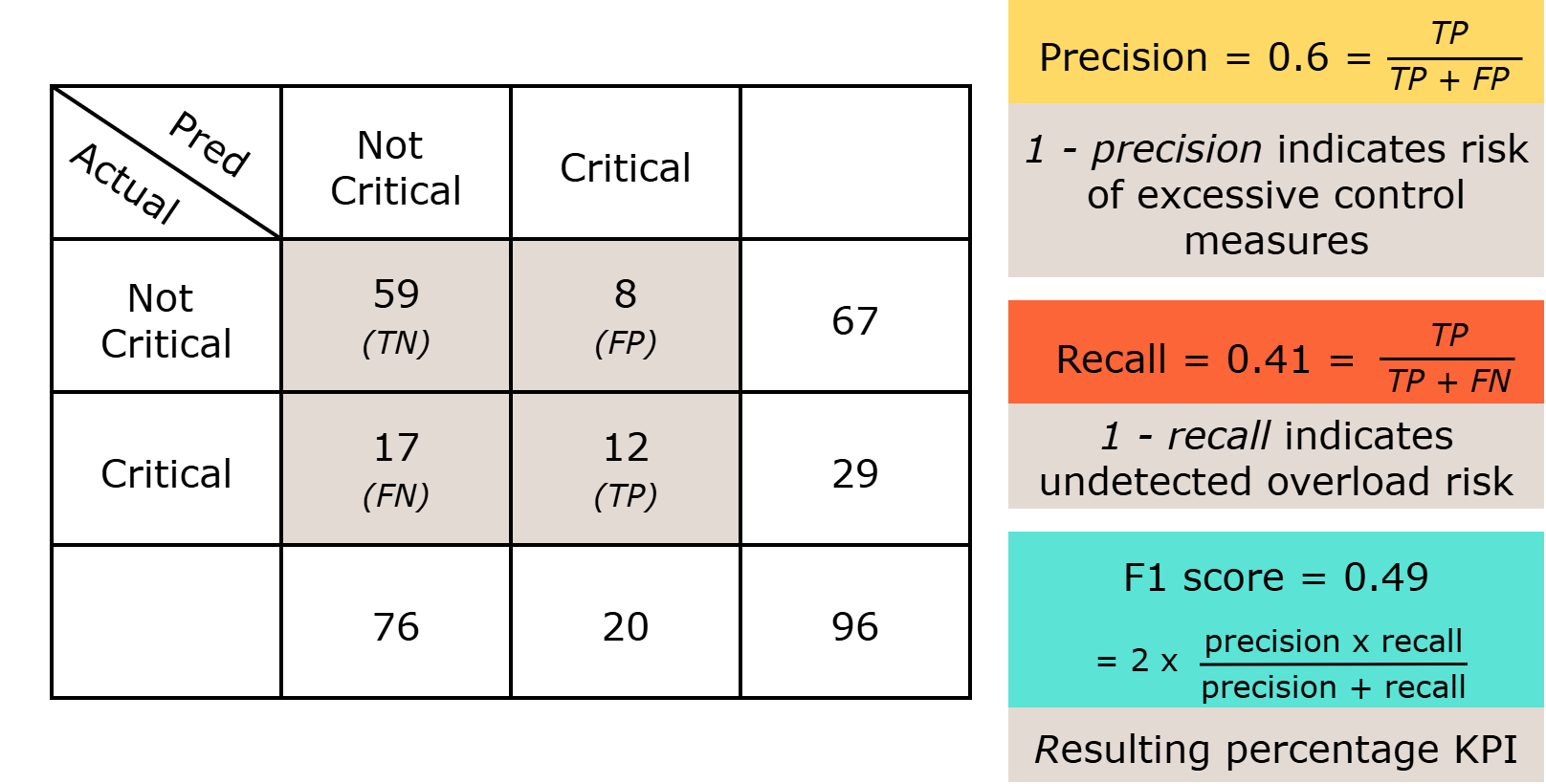}
        \caption{\textbf{Calculate metrics} resulting in a confusion matrix to derive final \ac{KPI}.}
        \label{subfig:metric_visualization_application_oriented_4}
    \end{subfigure}
    \caption{Visualization of the four steps to compute the application-oriented metric. The metric derives the parameters \textit{threshold} and \textit{window} from the fuse size of \ac{LV} feeders (\Cref{subfig:metric_visualization_application_oriented_1}), calculates the rolling mean independently for the actual ground truth and the prediction (\Cref{subfig:metric_visualization_application_oriented_2}), compares the actual ground truth to the prediction (\Cref{subfig:metric_visualization_application_oriented_3}), and results in a confusion matrix with percentage \ac{KPI} which is easy to communicate (\Cref{subfig:metric_visualization_application_oriented_4}). Prediction is abbreviated with \textit{pred}. Time-current curve is roughly redrawn from \cite{heuckElektrischeEnergieversorgungErzeugung2013}.}
    \label{fig:metric_visualization_application_oriented}
    \Description{Figure showing the calculation of the application-oriented metric in four steps.}
\end{figure*}

For \ac{LV} load forecasting evaluation, established metrics such as the \ac{MAE}, the \ac{RMSE} or R2-Score are important but lack practical relevance for \acp{DSO} when used alone. This is due to two reasons. First, these metrics do not focus on critical situations, in particular high loads that rarely occur, and second, they are difficult to communicate to people outside the machine learning field, particularly in a business context. Hence, we propose a novel application-oriented metric for \ac{LV} load forecasting that is tailored to \acp{DSO}, who aim to prevent overloading of grid assets such as \ac{LV} feeders while minimizing costs from curtailment measures and flexibility options. For this purpose, we map the core problem of \ac{LV} feeder overloads into its equivalent in the field of data science, resulting in a percentage \ac{KPI} that represents the harmonic mean of costs and risk of failure. Forecasts incur costs if control measures are conducted due to the forecast predicting high values, while a risk of failure, in the form of an overload, emerges if the forecast is lower than the actual values. The novel metric computation is divided into four steps, visualized in \Cref{fig:metric_visualization_application_oriented}.

\textbf{Derive parameters}: In the first step, two parameters, namely the \textit{window size} and the \textit{threshold}, must be chosen. These parameters are inspired by the time-current curve of a fuse used to protect the \ac{LV} feeder cable or line against overload. As shown in \Cref{subfig:metric_visualization_application_oriented_1}, the window size corresponds to the melting time and the threshold to the continuous short-circuit current of the time-current curve of the fuse. Two examples are illustrated, one for one minute (60 seconds) and one for one hour (3600 seconds). In practice, a pair of melting time and short-circuit current would specify a critical overload. However, in the data context, the threshold must be set much lower because values exceeding the rated current of the fuse are very rare and would therefore not allow for a proper evaluation. If the target variable is the three-phase active power as depicted in \Cref{fig:metric_visualization_application_oriented}, we can assume that the active power threshold $t_P$ corresponds to the current threshold $t_C$ with $t_P = 3 \cdot 230~\text{V} \cdot t_C$. Regarding the window size, the temporal resolution of the data must also be considered, as the window size must be a multiple of the temporal resolution. 

\textbf{Detect critical points}: In the second step, the actual ground truth value and the predicted (forecasted) values are considered separately, as depicted in \Cref{subfig:metric_visualization_application_oriented_2}. For both, the rolling mean is calculated according to the window size and compared to the threshold, with areas exceeding the threshold marked in blue. The predicted value can be the mean prediction but also any other quantile.

\textbf{Compare critical points}: The third step is shown in \Cref{subfig:metric_visualization_application_oriented_3}, where the rolling mean of the actual values and the predicted values are compared against each other point by point. Every point is classified into one of four categories: (1) true negative if both the actual and predicted values are below the threshold, (2) false negative if the actual value is above the threshold and the predicted value is below, (3) true positive if both the actual and predicted values are above the threshold, and (4) false positive if the predicted value is above the threshold while the actual value is below. 

\textbf{Calculate metrics}: Finally, the fourth step builds the confusion matrix and computes the precision, recall, and F1 score, which are common metrics in classification problems. \Cref{subfig:metric_visualization_application_oriented_4} shows the confusion matrix of the $96$ values for the exemplary day. Since there are more false negatives than false positives, the recall is lower compared to the precision in this example. Going back from the data science metric to the real-world problem, we can argue that $1 - precision$ is useful to assess the risk of overreaction when using the forecasting model in operation. On the contrary, $1 - recall$ represents the risk of taking insufficient action when using the model. The F1 score shows the overall performance of the forecast model, and its percentage value can be used to determine the forecasting quality that should be achieved. When evaluating long-term load forecasts for planning purposes with a forecast horizon of several years, $1 - \text{precision}$ corresponds to the risk of overinvestment and $1 - \text{recall}$ to the risk of underinvestment. 

\textbf{Peak metric variant}: Steps three and four can be omitted in favor of directly evaluating standard time series metrics on the subset of selected peak timestamps $T_{peak}$ based on the ground truth in step two. Hence, we define $T_{peak} \subseteq T_{all}$ as a subset of the set of all timestamps $T_{all}$ comprising all timestamps where the rolling mean of the ground truth exceeds the threshold $t$. The standard time series metrics such as \ac{MAE} or pinball loss only take into account the actual values and predictions of values in $T_{peak}$. Contrary to the proposed classification metrics, only the ground truth peaks are used for timestamp selection here to ensure comparability through equal peak timestamp counts across different model predictions. We note that for this variant, the standard time series metric is calculated on the original target values and predictions and not on the rolling means. 

The metric calculation described in steps one through four can be repeated for multiple threshold-window size pairs and their results subsequently combined. For each threshold-window size pair, we recommend to calculate two distinct confusion metrics for consumer (positive values) and producer (negative values) because this has the advantage of differentiating the forecasting quality regarding load and generation, and upper quantiles for consumption and lower quantiles for feed-in can be utilized as appropriate predictions. In general, we note that while the example in \Cref{fig:metric_visualization_application_oriented} only shows the example of one day, the metric is calculated for complete \ac{LV} feeders or the complete set of \ac{LV} feeders. Furthermore, it is possible to compute the metric for load estimations of other grid assets from \acp{DSO}. 

\section{Experimental Setup}\label{sec:experimental_setup}

\subsection{Dataset}
The dataset used for the experiments \ifanonymize{is}{is named \textit{FeederBW} and is} publicly available~\cite{treutleinRealworldEnergyData2026}. It comprises $200$ \ac{LV} feeders measured over two years and includes different covariates such as metadata about the \ac{LV} feeders and weather data. From the different measurements available at the \ac{LV} feeder, we select the active power over all three phases. The temporal resolution is changed from one minute to $15$ minutes data by mean aggregation since this temporal resolution is normally sufficient for processes at \acp{DSO}. \Cref{fig:distribution_static_columns_per_feeder} in the appendix and \Cref{fig:distribution_y_true} in the appendix show different histograms from the six metadata columns of the \ac{LV} feeders and the distribution of the $15$ minute values, which are used as target variable. On average, $36.2$ housing units and $2.5$ units of industry and commerce are connected by the \ac{LV} feeder. The target data is characterized by both consumption and generation depending on the concrete \ac{LV} feeder. We note that the weather data consists of short-term forecasts between one and three hours. Hence, our forecasts with a four-day forecast horizon are classified as ex-post forecast because in a real application, this weather data is not fully available at the forecast start~\cite{habenReviewLowVoltage2021}. 

\subsection{Models}
In addition to the three \acp{TSFM}, \Cref{tab:table_experimental_setup_models} comprises also six models that are fully trained from scratch as a benchmark and specifies the compute used for the models, the features provided to the models, and the application as zero-shot or fully trained. Further configurations and hyperparameters are specified in \Cref{tab:model_config_rest}. The first benchmark model \textit{WeekNaive} takes the average and quantiles of the same weekday over the last four weeks. All models predict a point estimate (mean or median), a lower $0.05$ quantile, and an upper $0.95$ quantile. Hence, the point prediction of WeekNaive corresponds to the mean and the lower and upper quantile to the minimum and maximum of the same weekday over the last four weeks. Two tree-based XGBoost models are included, one of which is purely based on features without context information of the time series itself, as proposed in \cite{treutleinGeneratingPeakawarePseudomeasurements2025}, while the second XGBoost model (XGBoost+) includes historical context information from the last four days. The context of XGBoost+ is less compared to other models due to runtime constraints. Three neural network benchmark models are used, namely an \ac{MLP} implemented in PyTorch as well as \ac{TFT} and PatchTST from AutoGluon. Finally, Chronos-Bolt, \mbox{Chronos-2}, and TabPFN-TS are used as \acp{TSFM}, where the covariate regressor used for Chronos-Bolt is CatBoost. The context window of $1344$ values corresponds to two weeks. 

\subsection{Metrics}

Several metrics for both point estimates and quantile predictions are used to evaluate all observations. For detailed definitions we refer to the literature, for example, the metrics presented in \cite{gonzalez-sopenaOverviewPerformanceEvaluation2021} in the context of short-term wind power forecasting. For the standard metrics, the metric result is calculated for every observation and prediction of a \ac{LV} feeder and then averaged for each individual feeder. Regarding point estimates, we use MAE and RMSE. Additionally, the R2 score is used which indicates the proportion of the variance is explained by the forecast. Regarding quantile predictions, pinball loss and Winkler score are used. The pinball loss is computed as a combined pinball loss for the $0.05$ and $0.95$ quantile without the point estimation. The Winkler score~\cite{Winkler01031972} evaluates an interval prediction with a lower and upper quantile $(\hat{y}_l, \hat{y}_u)$ and is defined as 
$$
WIS_{\alpha} = 
    \begin{cases}
       (\hat{y}_u - \hat{y}_l) + \frac{2}{\alpha}(\hat{y}_l - y) & \text{if } y < \hat{y}_l\\
       (\hat{y}_u - \hat{y}_l) & \text{if } \hat{y}_l \leq y \leq \hat{y}_u\\
       (\hat{y}_u - \hat{y}_l) + \frac{2}{\alpha}(y - \hat{y}_u) & \text{if } y > \hat{y}_u
    \end{cases}
$$
where $y$ represents the actual observation and $\alpha = 1 - coverage = 0.1$. The Winkler score is rewarding sharp intervals (across all cases, particularly when $\hat{y}_l \leq y \leq \hat{y}_u$) while penalizing observations outside the predicted interval (cases $y < \hat{y}_l$ and $y > \hat{y}_u$). Additionally, the interval width and the empirical coverage are computed. The optimal empirical coverage is $0.9$, corresponding to the $90$~\% interval between the quantiles $0.05$ and $0.95$.

For the application-oriented metric resulting in a confusion matrix, we use a window size of one hour leading to a rolling mean of four values due to the $15$-minutes resolution of the target variable. The threshold is chosen as $40$~\% of the absolute maximum of an individual feeder. Hence, the threshold differs between \ac{LV} feeders. As suggested in \Cref{subsec:calculation_of_application_oriented_metric}, we differentiate between consumer and producer peaks. Consequently, there are two resulting \acp{KPI} for both consumer and producer, each with the standard weighting of the F1 score (harmonic mean) between precision and recall. 

The MAE, pinball loss, Winkler score, interval width, and empirical coverage are also used for the peak metric variant introduced in \Cref{subsec:calculation_of_application_oriented_metric}. The difference compared to the standard metrics is that they are only calculated for the timestamps $T_{peak}$ where a peak is identified in the actual observation. The parameters for window size and threshold are the same as those used for the application-oriented metric resulting in a confusion matrix and we also differentiate between consumer and producer. 

\input{tables/table_experimental_setup_models}

\subsection{Procedure}

The models are trained and evaluated according to the split illustrated in \Cref{fig:train_test_split} in the appendix. The period from April 1, 2023 to March 31, 2024 serves as training data for the fully trained models across \ac{LV} feeders $1$--$160$. This period is excluded for the WeekNaive baseline and the \acp{TSFM}, as is the first year of data for feeders $161$--$200$. All models are evaluated on the second year from April 1, 2024 to March 31, 2025. The metric computation is conducted for every individual feeder separately and then aggregated to overall metric results. Forecasts are issued every four days for the next four days (forecast horizon), meaning that both the forecast horizon and the stride are four days. XGBoost+ is the only model that introduces a gap between the last timestamp of the past context and the first timestamp of the forecast horizon, where 12 steps or 3 hours are discarded. XGBoost+ forecasts autoregressively, where we observed that introducing a gap by masking the most current past time steps improved forecasting quality. Regarding the point forecast targets, WeekNaive uses the mean of the same day over the previous four weeks, the AutoGluon models use the mean output, TabPFN-TS is configured to produce mean forecasts, and the XGBoost variants as well as TorchMLP are set to forecast the median.

\section{Results}\label{sec:results}

\input{tables/table_metric_results_standard}

\subsection{Evaluation Overview}

\Cref{tab:metric_results_standard} summarizes the respective metrics for the six baseline models and the three \acp{TSFM}. The results of \acp{TSFM} without covariates are included in \Cref{subsec:covariates_ablation_study}. The average of a metric is calculated for each \ac{LV} feeder and then, for each metric, the median across the $200$ feeders is taken.

Regarding the point estimation metrics, for example the \ac{MAE}, \mbox{Chronos-2} is the best performing model with an \ac{MAE} of $3.839$~\si{\kW}, the second best model, TabPFN-TS, with an error of $4.137$~\si{\kW}. Similarly, \mbox{Chronos-2} outperforms the second best model, Chronos-Bolt, regarding the \ac{RMSE} and R\textsuperscript{2} score. The best performing baseline with respect to the point estimation metrics is the XGBoost+ model using the lagged features of the time series. However, all three \acp{TSFM} exhibit better values in all three metrics compared to the XGBoost+ model. In contrast to \mbox{Chronos-2}, the metric results of Chronos-Bolt and TabPFN-TS are close to the XGBoost+ model.

The \acp{TSFM} also perform better in terms of quantile metrics. The combined pinball loss for the $0.05$ and $0.95$ quantiles is $0.5545$~\si{\kW} for \mbox{Chronos-2}, followed by TabPFN-TS with $0.5767$~\si{\kW}, while the best baseline is again XGBoost+ with $0.6268$~\si{\kW}. The pinball loss of Chronos-Bolt is significantly higher at $0.8805$~\si{\kW}. The Winkler score shows the same ratio between models, with \mbox{Chronos-2} followed by TabPFN-TS, XGBoost+ as the best baseline, while Chronos-Bolt underperforms. Considering the interval width together with the empirical coverage indicates that Chronos-Bolt produces sharp intervals of $8.839$~\si{\kW} that miss many observations, as underlined by an empirical coverage of $62.38$~\% against an expected $90$~\%. While the WeekNaive also shows a low interval width of $11.58$~\si{\kW} combined with a low empirical coverage of $58.1$~\%, the other five baselines and two \acp{TSFM} are much closer to the expected $90$~\% empirical coverage, with results between $80.9$--$92.69$~\% and interval widths between $13.84$--$19.34$~\si{\kW}. Again, \mbox{Chronos-2} is closest to the optimum with an empirical coverage of $89.75$~\% and an interval width of $16.33$~\si{\kW}, which is in the middle compared to the other models.

\Cref{tab:runtimes} in the appendix provides the observed runtimes for training and inference of the models. The \acp{TSFM} and WeekNaive do not exhibit training time since they are applied in a zero-shot manner. An exception is Chronos-Bolt, where no \ac{TSFM} is trained but rather a CatBoost model to incorporate the covariates. Training times for the remaining models range between four minutes for PatchTST and more than four hours for XGBoost+. The inference time of the baselines is below four minutes, except for the XGBoost models, which take 15 minutes for XGBoost and more than two hours for XGBoost+. The inference times of \acp{TSFM} differ significantly, with Chronos-2 achieving fast inference in less than 25 minutes, while TabPFN-TS requires more than eight hours.

A qualitative result of the \mbox{Chronos-2} forecast is depicted in \Cref{fig:sketch_forecast_quantiles_may_2024_141_105} for feeder $F_{141}$, combining three forecasts with a forecast horizon of four days from May 1, 2024 to May 12, 2024. In the ground truth, some days exhibit significant feed-in up to $-50$~\si{\kW} measured at the \ac{LV} feeder while other days show virtually no reverse flow into the medium-voltage grid. Load spikes are lower in terms of magnitude and duration and occur primarily in the morning and evening. The point estimation reflects the basic course of the ground truth. However, some consumption-induced peaks and feed-in are not adequately represented in the point estimation, for example, the load in the morning of May 4 or the feed-in at noon of May 5. Nevertheless, these peak periods are adequately captured by the lower and upper quantiles. The interval width varies considerably, with tight intervals during the night and wide intervals at noon. Furthermore, the tight and low interval prediction at noon on May 9 with relatively little feed-in over the previous days demonstrates that the model adapts intervals to exogenous feature information in addition to time-related features.

\begin{figure*}
    \begin{subfigure}[t]{0.95\textwidth}
        \centering
        \includegraphics[width=1\linewidth]{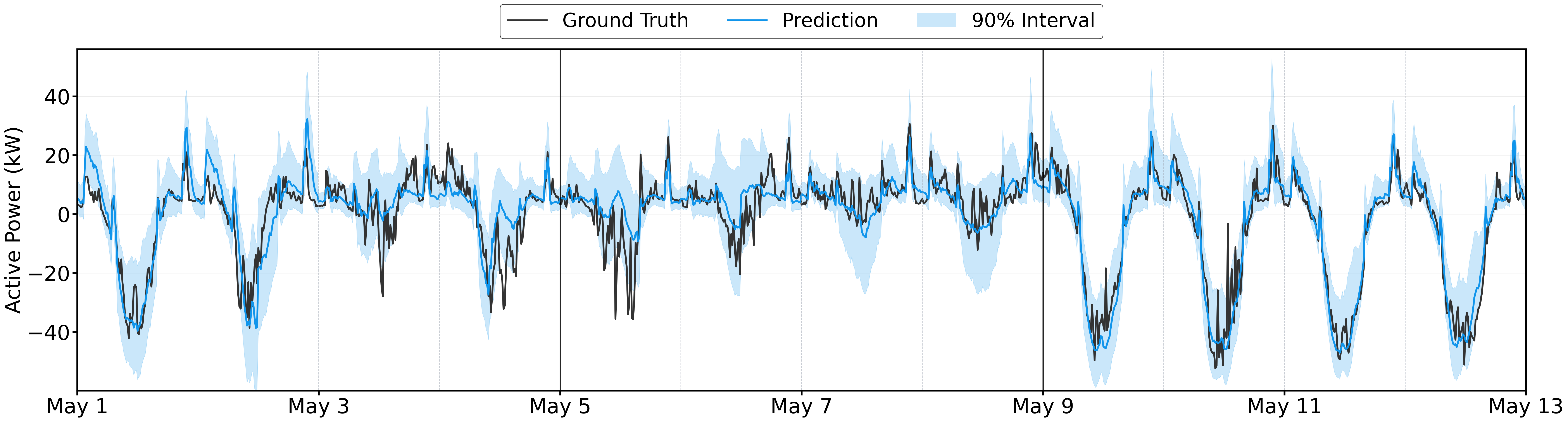}
        \caption{Covariate-informed including weather}
        \label{fig:sketch_forecast_quantiles_may_2024_141_105}
    \end{subfigure}
    \begin{subfigure}[t]{0.95\textwidth}
        \centering
        \includegraphics[width=1\linewidth]{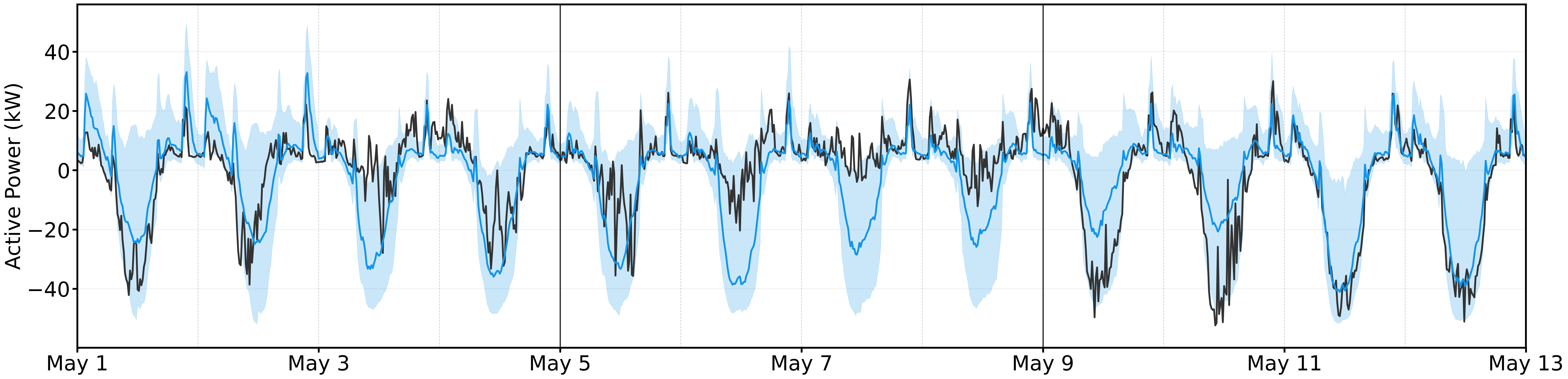}
        \caption{Univariate solely based on features of lagged time series}
        \label{fig:sketch_forecast_quantiles_may_2024_141_106}
    \end{subfigure}
    \caption{Excerpt of three \ac{LV} load forecasts of feeder $F_{141}$ from \mbox{Chronos-2} with a forecast horizon of four days from May 1, 2024 to May 12, 2024 (UTC). Note the wider intervals produced by the univariate model and the model predicting significant feed-in on all days, while the  weather-influenced forecast exhibits narrower intervals and is able to capture the time series better.}
    \label{fig:forecast_exemplary_excerpt_feeder_141}
    \Description{Excerpt of load forecast of \mbox{Chronos-2} with and without covariates.}
\end{figure*}

\Cref{fig:sketch_error_by_lead_time_105} depicts the forecast error (Winkler score) of \mbox{Chronos-2} for each of the $384$ individual forecasts within the four-day forecast horizon, aggregated over all forecasts in the test period across all $200$ feeders. Since the stride is four days, predictions starting at midnight (UTC) are always shifted by four days. The lines represent the mean error as well as the $0.25$, median and $0.75$ quantiles over all forecast errors at each time step within the forecast horizon. Notably, the error is comparably high at noon and low at night for all four days in the forecast horizon, which can be observed both in the mean and across all quantiles. The error band between the $0.25$ and $0.75$ quantiles is narrower during the night and wider around noon. Furthermore, the mean is close to the $0.75$ quantile, indicating that the error distribution at each time step within the forecast horizon is skewed.

\begin{figure*}
    \begin{subfigure}[t]{0.95\textwidth}
    \centering
        \includegraphics[width=1\linewidth]{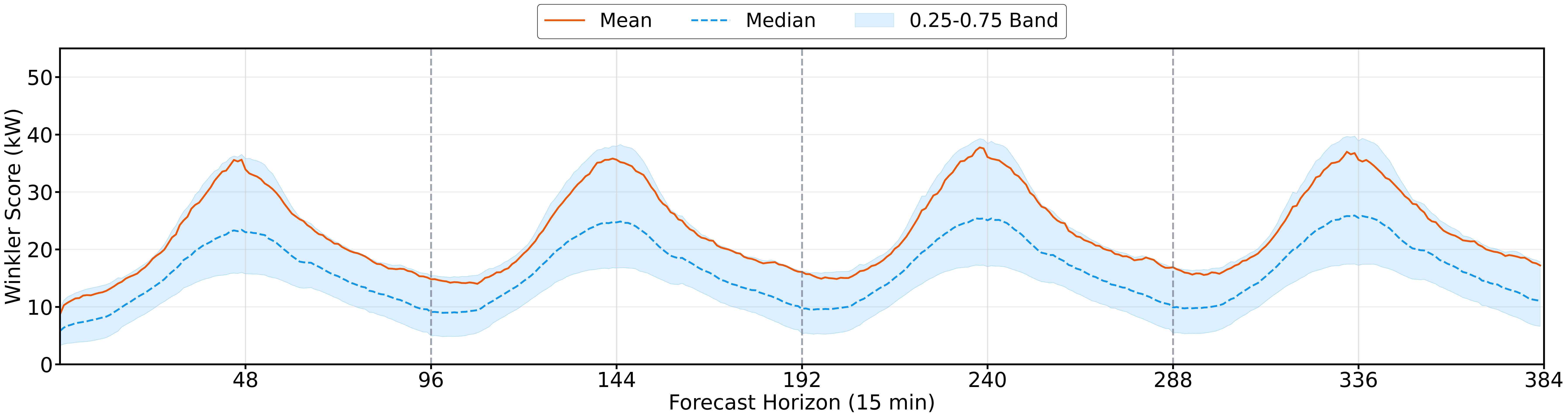}
        \caption{Covariate-informed including weather}
        \label{fig:sketch_error_by_lead_time_105}
    \end{subfigure}
    \begin{subfigure}[t]{0.95\textwidth}
        \includegraphics[width=1\linewidth]{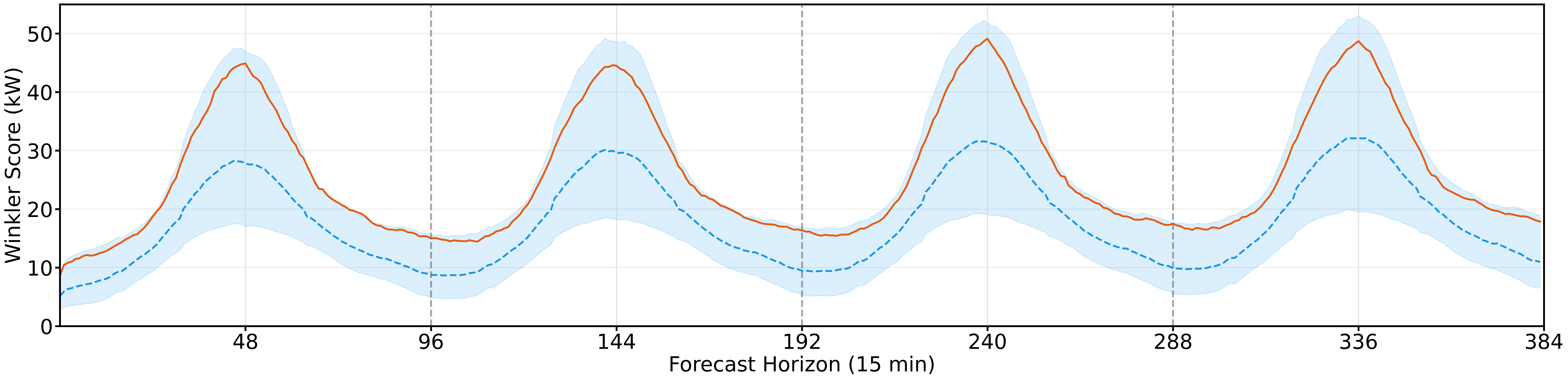}
        \caption{Univariate solely based on features of lagged time series}
        \label{fig:sketch_error_by_lead_time_106}
    \end{subfigure}
    \caption{Distribution of the Winkler score for \mbox{Chronos-2} over the forecast horizon. All forecasts over all \ac{LV} feeders are aggregated by their forecast horizon. The vertical lines indicate full days from midnight UTC, the start of all forecasts. The Winkler score distribution barely differs between the configurations during the nights, while the score values during the day are often much higher for the univariate model.}
    \Description{Error plot with respect to the forecast horizon of \mbox{Chronos-2}.}
\end{figure*}

\subsection{Covariates Ablation Study}\label{subsec:covariates_ablation_study}

\Cref{tab:metric_results_standard} presents for each \acp{TSFM} the metric results for inference without covariates, notably without weather information during the forecast horizon. Regarding the metrics evaluating the point estimation, all three \acp{TSFM} deteriorate to a similar extent when omitting covariates. The MAE rises by about $1$~\si{\kW} for \mbox{Chronos-2} from $3.839$~\si{\kW} to $4.813$~\si{\kW}. Analogously, the MAE of Chronos-Bolt increases from $4.16$~\si{\kW} to $5.118$~\si{\kW} and for TabPFN-TS from $4.137$~\si{\kW} to $4.996$~\si{\kW}. \mbox{Chronos-2} remains the best performing model among the three \acp{TSFM} with respect to the point forecast metrics. A deterioration can also be observed for the metrics evaluating the quantiles. The combined pinball loss increases from $0.5545$~\si{\kW} to $0.6193$~\si{\kW} for \mbox{Chronos-2}, from $0.5767$~\si{\kW} to $0.6726$~\si{\kW} for TabPFN-TS and from $0.8805$~\si{\kW} to $1.044$~\si{\kW} for Chronos-Bolt. Furthermore, the prediction intervals are wider for all three \acp{TSFM}, increasing between $1.5$--$1.961$~\si{\kW}, while the empirical coverage for Chronos-Bolt increases towards $90$~\% and the empirical coverage of TabPFN-TS decreases towards $90$~\%. Univariate \mbox{Chronos-2} worsens compared to the empirical coverage of the model incorporating covariates.

\Cref{fig:sketch_forecast_quantiles_may_2024_141_106} shows the same excerpt of \ac{LV} feeder $F_{141}$ for \mbox{Chronos-2} incorporating covariates in contrast to \Cref{fig:sketch_forecast_quantiles_may_2024_141_105}. It can be observed that excluding the covariates increases the interval width during noon for that period, where in particular the lower quantile exhibits lower values. The point estimation during noon is higher for several days with significant reverse power flow, such as May 1, May 2, and May 9. For other days, such as May 6--8 with nearly no feed-in, the point estimation is much lower and resembles a typical \ac{PV} curve. The change of the point estimation at the rest of the day is minor.

\Cref{fig:sketch_error_by_lead_time_106} visualizes the Winkler score for all feeders for \mbox{Chronos-2} without covariates. Higher errors compared to the same figure for \mbox{Chronos-2} with covariates in \Cref{fig:sketch_error_by_lead_time_105} are clearly visible during four periods for all quantiles and the mean. These four periods correspond to noon. Compared to the middle of the day, the error during the night, in the morning, and in the evening is not noticeably different between \mbox{Chronos-2} with and without covariates, at least upon visual inspection. The forecast error at day four (288 - 383) is visibly higher for the quantiles and the mean compared to day one.

\subsection{Application-Oriented Metric}\label{subsec:application_oriented_metric}

\input{tables/table_metric_results_application_oriented}

\Cref{tab:metric_results_application_oriented} shows the results of the novel application-oriented metric introduced in the present paper (see \Cref{subsec:calculation_of_application_oriented_metric}). For each model, we first differentiate between the consumer and producer perspectives. On the consumer side, critical peaks are identified in the ground truth or predicted values when values exceed the positive threshold. On the producer side, critical peaks are identified for values falling below the negative threshold. For both perspectives, precision, recall, and F1 score are calculated regarding different predicted values. The point variant compares against the mean or median, depending on the model, to calculate the metrics based on the point estimation. The quantile variant bases the calculation on the $0.95$ quantile for the consumer perspective and the $0.05$ quantile for the producer perspective.

Similarly to the other metrics, \acp{TSFM} achieve better results in the classification metrics compared to the baselines. An exception is the consumer precision (point), where TorchMLP achieves the highest value among all nine models while its consumer recall (point) is the second lowest. None of the \acp{TSFM} dominates across all classification metrics. However, we can observe a pattern that Chronos-Bolt achieves the best performance for the consumer and producer precision and F1 score in the quantile variant. \mbox{Chronos-2} exhibits the best values for the consumer and producer recall and F1 score in the point variant. TabPFN-TS achieves the best recall values in the quantile variant for both perspectives. We also note that WeekNaive is the second-best model regarding the consumer precision and F1 score in the quantile variant. Regarding the baselines on the producer side, XGBoost, which relies solely on metadata without lagged features of the time series, is second best for producer precision (point). XGBoost+ is second best for producer precision and F1 score in the quantile variant.

When comparing the metrics in \Cref{tab:metric_results_application_oriented} between the consumer and producer perspectives based on the best model for each metric, the values for producer peaks are higher than for consumer peaks across precision, recall, and F1 score. An exception is the consumer precision (point), where TorchMLP exhibits the highest value of $0.8833$ across all models. Regarding the F1 score, the best consumer model achieves $0.7496$ compared to $0.7838$ for the producer side (both \mbox{Chronos-2}) in the point variant, and $0.6678$ compared to $0.7379$ (both Chronos-Bolt) in the quantile variant. XGBoost, based solely on metadata, has the lowest consumer F1 score across all models with $0.5249$ in the point variant and $0.4787$ in the quantile variant. In contrast, the producer F1 score of XGBoost reaches $0.6091$ (point) and $0.625$ (quantile), resulting in a rank of 6 for the point variant and rank 3 for the quantile variant across all models. For WeekNaive, the consumer F1 score is comparably high with $0.6875$ in the point variant and $0.6283$ in the quantile variant, corresponding to rank 5 and rank 2 across all models. In contrast, the producer F1 score of WeekNaive is comparably low with $0.4118$ (point) and $0.5363$ (quantile), resulting in rank 8 and rank 6 across all models.

For all models, the precision in the point variant exhibits higher values compared to the quantile variant, while the recall values are higher for the quantile variant compared to the point variant for both the consumer and producer perspectives. Despite this trade-off between precision and recall, the F1 score of the best model is higher for the point variant compared to the quantile variant for both consumer and producer.

In addition to the peak classification metrics, \Cref{tab:metric_results_peakmetric} in the appendix provides metric results that are typically used in time series evaluation (MAE, pinball loss) as well as the interval width and empirical coverage, which are all only evaluated if the ground truth exhibits a peak. The peak detection follows the same procedure described in \Cref{subfig:metric_visualization_application_oriented_1} and \Cref{subfig:metric_visualization_application_oriented_2}, again divided into consumer and producer peaks. \Acp{TSFM} show superior performance compared to the baselines. The MAE and pinball loss evaluated only for ground truth peaks are higher compared to the metrics evaluated for the complete time series, as depicted in \Cref{tab:metric_results_standard}. While the median MAE is $3.839$~\si{\kW} for \mbox{Chronos-2} for the full series, it is $7.304$~\si{\kW} when evaluating only in the presence of consumption ground truth peaks and $8.786$~\si{\kW} when evaluating only in the presence of producer ground truth peaks. The aggregated interval width per model is higher in the peak periods of the ground truth. For example, \mbox{Chronos-2} exhibits an interval width of $16.33$~\si{\kW} for the complete time series, while the interval width increases to $20.28$~\si{\kW} (consumer) and $33.76$~\si{\kW} (producer). All models exhibit a lower empirical coverage in the presence of ground truth peaks compared to the empirical coverage calculated for the complete time series. Comparing consumer and producer peaks, the MAE, pinball loss, and interval width of the respective best model are higher for producer peaks than for consumer peaks.

\section{Discussion}\label{sec:discussion}

\subsection{Overall Performance}

The evaluation across all examined metrics, including standard metrics evaluating the complete time series, metrics considering only ground truth peaks, and the proposed application-oriented metric, reveals that the best-performing model comes from the class of TSFMs. The only exception is the TorchMLP model for the consumer precision (point) metric. This indicates that TSFMs are well-suited for LV load forecasting and outperform existing models that were specifically trained on the dataset.

Looking at the individual TSFM performances in more detail, \mbox{Chronos-2} shows the best overall performance at the point estimation metrics and uncertainty quantification through quantiles while also achieving fast inference times. TabPFN-TS demonstrates comparable performance with respect to the metrics, but its inference times are much larger compared to \mbox{Chronos-2}. A possible reason for this could be TabPFN's focus on broader tabular data, and the ability to work on smaller datasets, compared to the \mbox{Chronos-2} model, which focuses exclusively on time-series forecasting and also supports even longer contexts. In contrast, Chronos-Bolt performs much worse for uncertainty prediction compared to mean prediction. The sharp and narrow intervals lead to a coverage that is only undercut by WeekNaive, thus showing an inadequate trade-off between interval width and empirical coverage.

Concerning the computational and temporal requirements, \Acp{TSFM} themselves need no training time, which is one of their main advantages. While Chronos-Bolt itself does not need task-specific training, its covariate regressor has to be fitted to be able to incorporate covariates.

Inference times of the examined models differ considerably. Chro\-nos-Bolt is relatively slow at 1.5 hours. The newer \mbox{Chronos-2} is reasonably fast at under half an hour, though still being beaten by most baselines time-wise. TabPFN-TS, however, is very slow, with the longest overall inference time of over eight hours. \mbox{Chronos-2}'s $81.6$~ms per forecast paired with its good performance could enable applications where inference is conducted more frequently.

The plots of the Winkler score distribution over the forecast horizon do not reveal the logarithmic development often seen for such error plots. Potential reasons why this typical behavior is not observed here are connected to the used covariates and dataset. In the \Ac{LV} grid in general and in this dataset specifically, there are many PV plants installed across the feeders, as seen in \Cref{fig:distribution_static_columns_per_feeder} in the appendix. These are expected to be highly influential on the measured active power. With PV plants having such a significant impact, weather data and its quality will induce additional error into the forecasts. The used weather data is at an hourly and not a 15-minute temporal resolution, and its spatial resolution is at postal code level rather than feeder level. Using shorter stride lengths that do not perfectly align with day and night might also lead to a more typical error behavior.

Furthermore, the used weather data, as included in the dataset, is of the same "quality" at every step. At day 4, we can rely on the same weather forecast quality compared to day 1 because it all stems from short-term forecasts of similar lead times (ex-post). The small changes that are seen in the error curves from day to day are heavily overshadowed by the large deviations in error between day and night.

Another apparent peculiarity of the Winkler score distribution plots in \Cref{fig:sketch_error_by_lead_time_105,fig:sketch_error_by_lead_time_106} is their skewness. The majority of observations apparently have lower errors since the median is significantly lower compared to the mean, while a smaller subset of observations exhibits high errors, skewing the distribution for each time point in the forecast horizon.

We can clearly confirm some findings from \cite{habenShortTermLoad2019}: The time of day plays an important role for accuracy but consistent with their observations, no large difference in error between day-ahead versus four-day-ahead forecasts is seen by Haben et al.~\cite{habenShortTermLoad2019} and in the present study.

While the examined models differ compared to Meyer et al. \cite{meyerBenchmarkingTimeSeries2025}, and their study does not incorporate covariates such as weather data, we similarly observe that \Acp{TSFM} outperforming the selected baselines. Moreover, the \Acp{TSFM} themselves show partly large differences in performance, which aligns with their findings.

Despite these promising results, several limitations of the present study should be acknowledged. First, there is one model (XGBoost+) with a different context window. Additionally, taking the median over the 200 feeders for the metric values may hide the individual feeder performance, though the boxplot over Winkler scores per feeders included in \Cref{fig:wis_boxplot} in the appendix does so. As the prevalence of technologies like \Ac{PV} and batteries varies between feeders, model performance will vary as well. Furthermore, a reduction of the stride from four days towards the temporal resolution of the data (15 minutes) to make more forecasts could provide more insights, with each data point in the dataset covered by multiple forecasts, as opposed to a single forecast per data point in the used configuration.

It should also be noted that more sophisticated non-TSFM approaches would likely be able to narrow the gap between \Acp{TSFM} and baselines. The performance statements made are not definitively saying that \Acp{TSFM} are better. However, the out-of-the-box performance of \Acp{TSFM} is very good. From a methodological perspective, inference times are only measured using single runs, and some deviation from run to run is to be expected. Finally, these results are limited to the examined models and the used data.

\subsection{Covariates}

As seen in the short ablation study, running the \Acp{TSFM} as univariate models by removing covariates significantly reduces performance. Nevertheless, the univariate \Acp{TSFM} are still relatively good, also compared to the covariate-informed benchmarks, which demonstrates their robustness.

For \acp{DSO}, weather covariates are helpful to reduce uncertainty regarding weather conditions, in particular for PV generation. Including the solar radiation significantly improves the quantile estimation during noon. Beyond uncertainty quantification, the covariates also improve the performance of the point estimation.

In addition, the error distribution is dampened by covariates in relation to the forecast horizon. The error around noon increases slightly more from day 1 to day 4 along the forecast horizon if the model is not informed by covariates. Consequently, the aforementioned positive effect of always being able to rely on the same quality of weather data does not exist for the univariate case. This means these models are more dependent on the past target time series at the end of the forecast horizon.

However, some limitations regarding the ablation study should be noted. While the performance of the univariate \Acp{TSFM} is impressive, they are not compared against univariate non-TSFMs, except WeekNaive. Also, for the ablation study, we only removed all covariates simultaneously and did not examine their individual impact. For example, statements about the importance of single covariates like the temperature, as made by \cite{habenShortTermLoad2019}, cannot be discerned from the other covariates this way.

\subsection{Application-Based Assessment}

The proposed application-oriented metric exhibits several advantages. First, it provides data scientists a clear indication of the forecasting performance with respect to consumption and production, as it links widely used classification metrics in the form of precision, recall and the F1 score to the domain problem of under- and over-dimensioning of \ac{LV} grid assets. Second, it is easy to communicate the resulting percentage \acp{KPI} to other stakeholders of the \ac{LV} load forecast who are not necessarily data scientists, such as electrical engineers or the management board. The interpretation of the forecasting quality in view of the domain requirements facilitates exchange and the establishment of targets for forecast quality. Third, it provides a way to determine the impact of integrating uncertainty through quantiles into the processes. \acp{DSO} can decide whether decisions are based on the point estimation or quantile predictions of the model, and the application-oriented metric provides information on the degree to which this will influence costs (precision) and the risk of failure (recall).

In classical peak metrics such as those presented in \Cref{tab:metric_results_peakmetric} in the appendix, the selection of peaks used for the metrics is based solely on the ground truth time series and not on the prediction. Therefore, a model that falsely predicts peaks where there are none is not penalized. However, such a forecasting model also incurs costs if planning and operation are aligned with the predictions. In the application-oriented metric based on classification metrics, such cases are reflected in the number of false positives. Consequently, classification metrics additionally reveal these aspects of forecast quality and are therefore considered a worthwhile extension. For both the peak metrics in \Cref{tab:metric_results_peakmetric} in the appendix and the application-oriented metric in \Cref{tab:metric_results_application_oriented}, we evaluate for an unlimited number of peaks. Hence, compared to peak metrics considering only daily peaks as proposed in~\cite{shuklaBigDEALChallenge20222024, treutleinGeneratingPeakawarePseudomeasurements2025}, we can account for the evaluation of several daily peaks, for example, in the morning and evening. Furthermore, \ac{LV} feeders characterized by industry and commerce may exhibit even more regular peaks.

The application-oriented metric clearly shows the trade-off between using point estimation and using quantiles as a basis for decision making. Choosing quantiles instead of point estimation results in decreased precision and increased recall for all models. This is expected because the quantile estimation is higher for consumer peaks or lower for producer peaks compared to the point estimation. Hence, a quantile predicts peaks more often, increasing true positives. While the increase in true positives reduces false negative predictions, the false positives will likely increase as well. Precision and recall can be weighted to compute the F1 score depending on the preferences of the \ac{DSO}. More cost-sensitive applications would aim to increase precision, while a focus on preventing asset overloads would favor a higher recall.

The peak metrics indicate that producer peaks are more complicated to estimate compared to consumer peaks. However, except for consumer precision (point), the application-oriented metrics for producer peaks are better compared to consumer peaks. A possible reason is that peak metrics using \ac{MAE} and pinball loss focus more on the performance regarding the exact value of the peak magnitude compared to the application-oriented metric, which considers only whether the prediction exceeds the threshold or not. Hence, predicting the time of producer peaks could be easier compared to consumer peaks, but the magnitude is more difficult. This would be reasonable since consumption data is often spiky due to the stochastic nature of electricity usage, while the producer peak can be expected around noon at the time with the highest solar radiation. 

It is worth noting that the difference in performance with respect to consumer peaks and producer peaks may also be a result of the dataset characteristics. The uncertainty of the models for producer peaks is highlighted by the high interval widths for producer peaks compared to consumer peaks. An important feature for predicting producer peaks is weather information. Improving its limited temporal resolution might hold potential for more accurate producer peak prediction. Moving from ex-post weather data to four-day weather forecasts made at the forecast origin (ex-ante) will likely also introduce significantly higher error.

Another difference between consumer and producer peaks is shown by the baseline models WeekNaive and XGBoost. WeekNaive is only based on the past time series and performs comparably well on consumer application-oriented metrics but worse regarding the producer perspective. In contrast, XGBoost, based only on feeder metadata and without lagged time series data, performs comparably worse for consumer application-oriented metrics and better with respect to producer peaks. This leads to the assumption that the past time series might be more important to predict consumption of \ac{LV} feeders compared to predicting the generation, which is expected to be, in comparison, more dependent on the current solar radiation than on the past features of the time series.

The proposed application-oriented metric comprises several parameters that require careful selection. First, the threshold for counting peaks and the window size for calculating the rolling mean must be chosen. Since the real measured data is much lower compared to the parameters directly derivable from the time-current curve of the fuse, lower values must be selected. A practical choice is provided with a window size of one hour and a threshold based on the min-max range of the \ac{LV} feeder. However, a more in-depth investigation of these parameters is meaningful, as they influence the resulting confusion matrix. 

Furthermore, predicting more quantiles and comparing the resulting application-oriented metrics could provide additional insights. In addition, different \ac{DSO} preferences with other weightings between precision and recall compared to the F1 score could be considered. In reality, it is not relevant whether an overload is induced by consumption or generation, which supports using a single threshold for both producer and consumer peaks. However, choosing separate thresholds for producer and consumer peaks could better reflect \ac{LV} feeders where the magnitude of consumption peaks and generation peaks differ significantly. This which would evaluate peaks better from a pure time series perspective but the metric would lose its real-world connection. A limitation of the classification is that it does not evaluate the magnitude of a predicted peak. If a prediction crosses the peak threshold, it is counted as a peak regardless of the extent of the excess.

\section{Conclusion}\label{sec:conclusion}

\Acf{LV} load forecasting plays a central role in planning and operating \ac{LV} grids, and \acfp{TSFM} open up new possibilities for the rapid implementation of forecasting pipelines since training on specific datasets is almost unnecessary. In the present study, three \acp{TSFM} and six baselines are evaluated on a dataset with $200$ real-world \ac{LV} feeders for predicting the three-phase active power. The metric results show that the \acp{TSFM} Chronos-Bolt, \mbox{Chronos-2} and TabPFN-TS largely outperform the baseline models, in particular \mbox{Chronos-2}. Furthermore, we show that \acp{TSFM} are capable of adapting to covariate input, which is in particular important for predicting \acf{PV} generation which depends on solar radiation. In addition, the present study focuses on evaluating forecasts with respect to the application at \acfp{DSO}, where an application-oriented metric links the common data science metrics precision, recall and F1 score to the challenge of over- and under-dimensioning of \ac{LV} grid assets. \acp{TSFM} promise several advantages compared to task- and dataset-specific models, such as the elimination of training time, code reduction in forecasting pipelines, and the support of probabilistic forecasts out of the box. In light of this, we recommend focusing future research for load forecasting of \ac{LV} feeders on \acp{TSFM}.

The comparison of different \acp{TSFM} presented in this work represents a snapshot in a rapidly evolving research environment. While we could observe the superiority of certain models for our task and dataset, ongoing evaluation of further forecasting tasks in \ac{LV} grids is required, incorporating the newest \acp{TSFM} and other datasets. Furthermore, proper covariate integration in \acp{TSFM} remains an interesting research field for \ac{LV} load forecasting. Instead of forecasting the 15-minute values, one could also forecast the rolling mean directly, which may be more aligned with \ac{DSO} use cases. Future work could additionally examine how to adapt the loss function of load forecasting models to optimize with respect to the proposed application-oriented metric, though we recommend keeping the use of quantile regression or other probabilistic approaches. Further studies could also evaluate the robustness of the proposed application-oriented metric for different parameters and other datasets.

\begin{acks}
Funding was provided by the Helmholtz Association through the 'Energy System Design' program and the Helmholtz Association's Initiative Helmholtz AI.
\end{acks}

\printbibliography

\appendix

\section{Time Series Foundation Models}\label{sec:time_series_foundation_models}

\subsection{Paradigm Change}
Foundation models are large-scale models trained on broad data frequently using self-supervision~\cite{bommasaniOpportunitiesRisksFoundation2021}. They are often characterized by emergence and homogenization, where emergence describes the development of unplanned behavioral attributes while homogenization describes the convergence toward the same few underlying models. Another important attribute of foundation models is prompt sensitivity, emphasizing the importance of prompt engineering as the main interface to influence model behavior~\cite{schneiderFoundationModelsNew2024}. The overall number of models needed for machine learning tasks is significantly reduced through emergence, homogenization, and prompt engineering. Rather than training task-specific models from scratch, they follow a \textit{pre-train and adapt} paradigm, offering significant cost savings and ease of use through fine-tuning and prompt engineering. The literature and debate about foundation models are largely influenced by \acp{LLM}. However, other foundation models are rapidly emerging, for example, for vision tasks~\cite{awaisFoundationModelsDefining2025}, the earth system~\cite{bodnarFoundationModelEarth2025}, tabular data~\cite{hollmannAccuratePredictionsSmall2025} or power flow calculation~\cite{hamannFoundationModelsElectric2024a}.

The development of \acp{TSFM} is largely motivated by the supposed structural similarity between time series and language. Both exhibit sequential dependencies across varying ranges and complexities, which makes transformer-based architectures a natural fit for transfer to the time series domain. Unlike \acp{LLM}, \acp{TSFM} are generally smaller with respect to parameters. In general, \acp{TSFM} represent a novel class of \ac{AI} models trained on diverse, multi-domain time series data to enable zero- or few-shot inference across unseen tasks and domains, in particular for time series forecasting, thereby eliminating or reducing the need for task-specific training. This significantly reduces the effort for typical \ac{ML} tasks. The historical time series and other covariates given to the \acp{TSFM} correspond to the language prompt for \acp{LLM}. Similar to \acp{LLM}, \acp{TSFM} can be used out of the box (zero-shot) without fine-tuning at all.

\subsection{Chronological Development}
For a comprehensive overview, \Cref{tab:overview_timeseries_foundation_models} shows an alphabetically ordered list of \acp{TSFM} including details about the organization, publication year, number of model parameters, and the architecture. The development of \acp{TSFM} began with efforts to repurpose existing \acp{LLM} for time series data, such as GPT4TS~\cite{zhouOneFitsAll2023} and LLM4TS~\cite{changLLM4TSAligningPreTrained2024}, which fine-tuned pre-trained \acp{LLM}, while others like Time-LLM~\cite{jinTimeLLMTimeSeries2024} kept the language model frozen and only adapted the input and output layers. The first models trained from scratch on time series data, such as TimeGPT~\cite{garzaTimeGPT12024}, TimesFM~\cite{dasDecoderonlyFoundationModel2024}, Lag-Llama~\cite{rasulLagLlamaFoundationModels2023}, and ForecastPFN~\cite{dooleyForecastPFNSyntheticallyTrainedZeroShot2023}, emerged around late 2023 and early 2024, establishing the foundation for the \ac{TSFM} model class. It should be noted that several non-pre-trained models provided key architectural components that were subsequently incorporated into many \acp{TSFM}, for example, PatchTST~\cite{nieTimeSeriesWorth2023} and TSMixer~\cite{ekambaramTSMixerLightweightMLPMixer2023a}. Subsequent models introduced new capabilities, including covariate support (Moirai~\cite{wooUnifiedTrainingUniversal2024}), alternative tokenization strategies (Chronos~\cite{ansariChronosLearningLanguage2024}), and multi-task learning (Moment~\cite{goswamiMOMENTFamilyOpen2024}, UniTS~\cite{gaoUniTSUnifiedMultiTask2024}), broadening the applicability of \acp{TSFM}. Later in 2024, the focus shifted toward efficiency and alternative architectures, with models such as TTM~\cite{ekambaramTinyTimeMixers2024}, RWKV-TS~\cite{houRWKVTSTraditionalRecurrent2024}, and Chronos-Bolt~\cite{abdulfatiransariFastAccurateZeroshot2024,ansariChronosLearningLanguage2024} offering improved speed and reduced hardware requirements, while mixture-of-experts approaches like Moirai-MoE~\cite{liuMoiraiMoEEmpoweringTime2024} and Time-MoE~\cite{shiTimeMoEBillionScaleTime2025} aimed to improve scalability. In 2025, further advances included improved covariate handling through cross-feature attention mechanisms, as seen in TabPFN-TS~\cite{hollmannTabPFNTransformerThat2023,hollmannAccuratePredictionsSmall2025,hooTablesTimeHow2025} and \mbox{Chronos-2}~\cite{ansariChronos2UnivariateUniversal2025}, as well as novel architectural choices such as flow-matching-based losses (Sundial~\cite{liuSundialFamilyHighly2025}) and ensemble-based prediction merging the results of several \acp{TSFM} (Synapse)~\cite{dasSynapseAdaptiveArbitration2025a}.

\input{tables/table_overview_timeseries_foundation_models}

\subsection{Architectures and Characteristics}

The review in \cite{liangFoundationModelsTime2024a} provides a comprehensive overview of \acp{TSFM} and their architectures. Furthermore, some shared characteristics between \acp{TSFM} are revealed in the model overview in \Cref{tab:overview_timeseries_foundation_models}. Both academia and private companies are actively developing \acp{TSFM}, which emphasizes the rapid transition from research to practice and the expectation of significant financial benefits, further underlined by the very recent publication years. The number of parameters is predominantly in the range of 800 thousand to $2.4$ billion which is about three orders of magnitude lower compared to \acp{LLM}~\cite{EpochAIModels2025}. From an architectural perspective, transformers are largely dominating~\cite{liangFoundationModelsTime2024a}. However, the superiority of an encoder-decoder, encoder-only, or decoder-only architecture remains ambiguous~\cite{liangFoundationModelsTime2024a}.

Pre-training and dataset selection are crucial for \acp{TSFM} to generalize across diverse time series domains. Many \acp{TSFM} use self-supervised learning~\cite{liangFoundationModelsTime2024a}, either through next-token prediction \cite{ansariChronosLearningLanguage2024,dasDecoderonlyFoundationModel2024,liuSundialFamilyHighly2025,liuTimerGenerativePretrained2024} or masked reconstruction~\cite{goswamiMOMENTFamilyOpen2024,wooUnifiedTrainingUniversal2024}, often combining real-world and synthetic data to enhance robustness. Pre-trained models can be adapted to new tasks using few-shot (minimal fine-tuning) and full-shot (extensive fine-tuning) approaches or directly applied in a zero-shot manner (no additional training) with sufficient context provided at inference time. \acp{TSFM} are generally trained globally across multiple time series, unlike local models trained separately for each time series. Regarding preprocessing, normalization and scaling methods, such as mean scaling~\cite{ansariChronosLearningLanguage2024} and reversible instance normalization~\cite{auerTiRexZeroShotForecasting2025,ekambaramTinyTimeMixers2024,goswamiMOMENTFamilyOpen2024}, are used to stabilize data ranges and support model transferability. Tokenization and segmentation strategies vary, with some models using discrete bins or patches to process time series data~\cite{ansariChronos2UnivariateUniversal2025}. Models also incorporate temporal context through features like lag indices, calendar covariates, and seasonality information extracted via Fourier transforms~\cite{hooTablesTimeHow2025,rasulLagLlamaFoundationModels2023}. 

In the first generation of \acp{TSFM}, \acp{LLM} such as GPT and Llama were adapted for time series forecasting by treating numbers as text or aligning embeddings. Currently, as shown in \Cref{tab:overview_timeseries_foundation_models}, transformer-based models trained exclusively on time series data are dominating. The \acp{TSFM} leverage self-attention for long-range dependencies and can be structured as encoder-only (e.g. Moirai~\cite{wooUnifiedTrainingUniversal2024}), decoder-only (e.g. TimesFM~\cite{dasDecoderonlyFoundationModel2024}), or encoder-decoder (e.g. Chronos-2~\cite{ansariChronos2UnivariateUniversal2025}) architectures, each with distinct forecasting strategies. Incorporating covariates is an ongoing research question in \acp{TSFM}. \mbox{Chronos-2}, TabPFN, and UniTS apply attention mechanisms across time and feature dimensions to propagate covariate information~\cite{ansariChronos2UnivariateUniversal2025, hooTablesTimeHow2025,gaoUniTSUnifiedMultiTask2024}, while Moirai concatenates covariates and targets into a univariate series for transformer processing~\cite{wooUnifiedTrainingUniversal2024}. Models lacking covariate support, such as Chronos(-Bolt), Toto, and Sundial, can use residual boosting with a base regressor and a \ac{TSFM}, requiring both past and future covariates, though residual boosting is unsuitable for multivariate targets due to unknown future values. In contrast, ChronosX employs input and output adapters for covariate integration~\cite{arangoChronosXAdaptingPretrained2025a}. 

In-context learning in \acp{TSFM} refers to extracting information from past data without modifying model weights, and it is an important capability to enable zero-shot forecasting~\cite{bommasaniOpportunitiesRisksFoundation2021}. While some architectures like Chronos~\cite{ansariChronosLearningLanguage2024} are limited in capturing cross-temporal covariate-target relationships, newer models like \mbox{Chronos-2} and TabPFN use two-dimensional attention mechanisms that can learn these diagonal dependencies across both time and feature dimensions from the provided context during inference~\cite{ansariChronos2UnivariateUniversal2025,hooTablesTimeHow2025}. Some models are also capable of estimating the uncertainty of their forecasts. Moirai and Toto predict parameters of complex distributions, for example Student's t or log-normal mixtures, and then sample point predictions and quantiles from these~\cite{cohenThisTimeDifferent2025,wooUnifiedTrainingUniversal2024}. Other \acp{TSFM} like TiRex and \mbox{Chronos-2} directly estimate a fixed grid of quantiles using specialized loss functions~\cite{auerTiRexZeroShotForecasting2025,ansariChronos2UnivariateUniversal2025}. Some approaches like TiRex treat future inputs as missing values to better propagate uncertainty across long horizons~\cite{auerTiRexZeroShotForecasting2025}.

\input{tables/table_hyperparameter_models}

\begin{figure}
    \centering
    \includegraphics[width=1\linewidth]{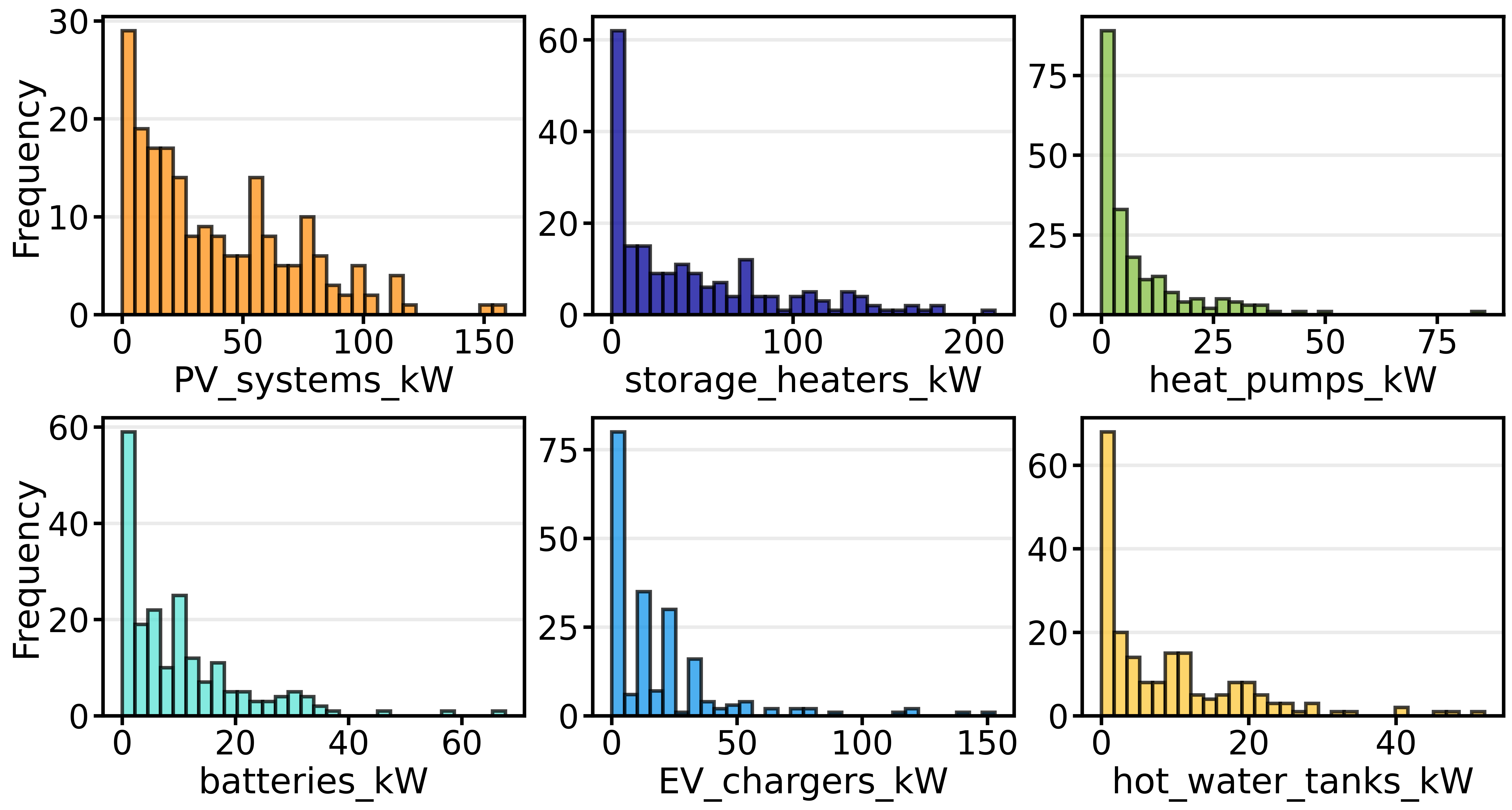}
    \caption{Frequency of six selected metadata columns with respect to all $200$ \ac{LV} feeders from the dataset used in our experiments~\cite{treutleinRealworldEnergyData2026}. The columns are aggregated per feeder using their maximum. The figure illustrates the diversity of the \ac{LV} feeders.}
    \label{fig:distribution_static_columns_per_feeder}
    \Description{Histograms of six selected metadata columns from the dataset.}
\end{figure}

\begin{figure}
    \centering
    \includegraphics[width=1\linewidth]{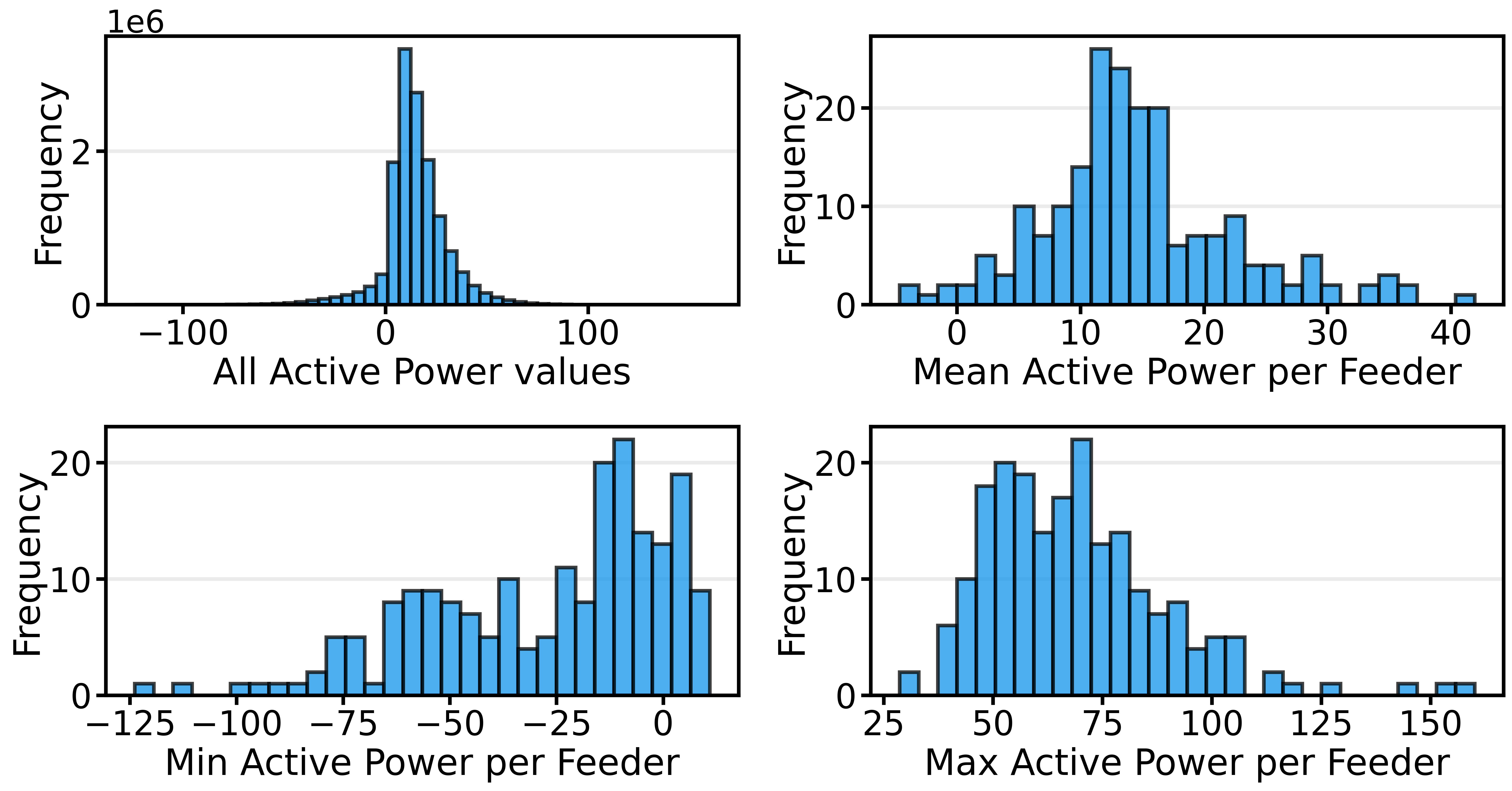}
    \caption{Active power with $15$ minutes mean aggregation which is used as target data from the dataset used in our experiments~\cite{treutleinRealworldEnergyData2026}. The histogram over all values is supplemented with histograms for the mean, minimum and maximum value per \ac{LV} feeder. The figure illustrates the diversity of the \ac{LV} feeders.}
    \label{fig:distribution_y_true}
    \Description{Histograms for mean, max, min and all values of target variable.}
\end{figure}

\begin{figure}
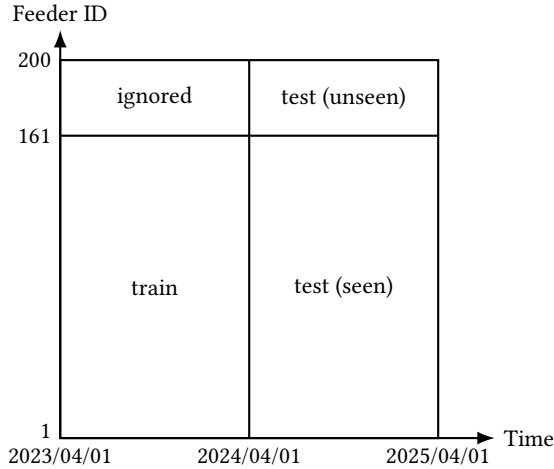

    \centering
    \include{tables/train_test_tikz.tex}
    \vspace{-2.5em}
    \caption{Train-test split used in the experiment for the models. While fully trained models use the period from April 1, 2023 to March 31, 2024 of the \ac{LV} feeders $1 - 160$ for training, \acp{TSFM} ignore this data.}
    \label{fig:train_test_split}
    \Description{Visualization of train-test split for the $200$ \ac{LV} feeders.}
\end{figure}

\input{tables/table_metric_results_peakmetrics}

\input{tables/table_runtimes}

\begin{figure}
    \centering
    \includegraphics[width=1\linewidth]{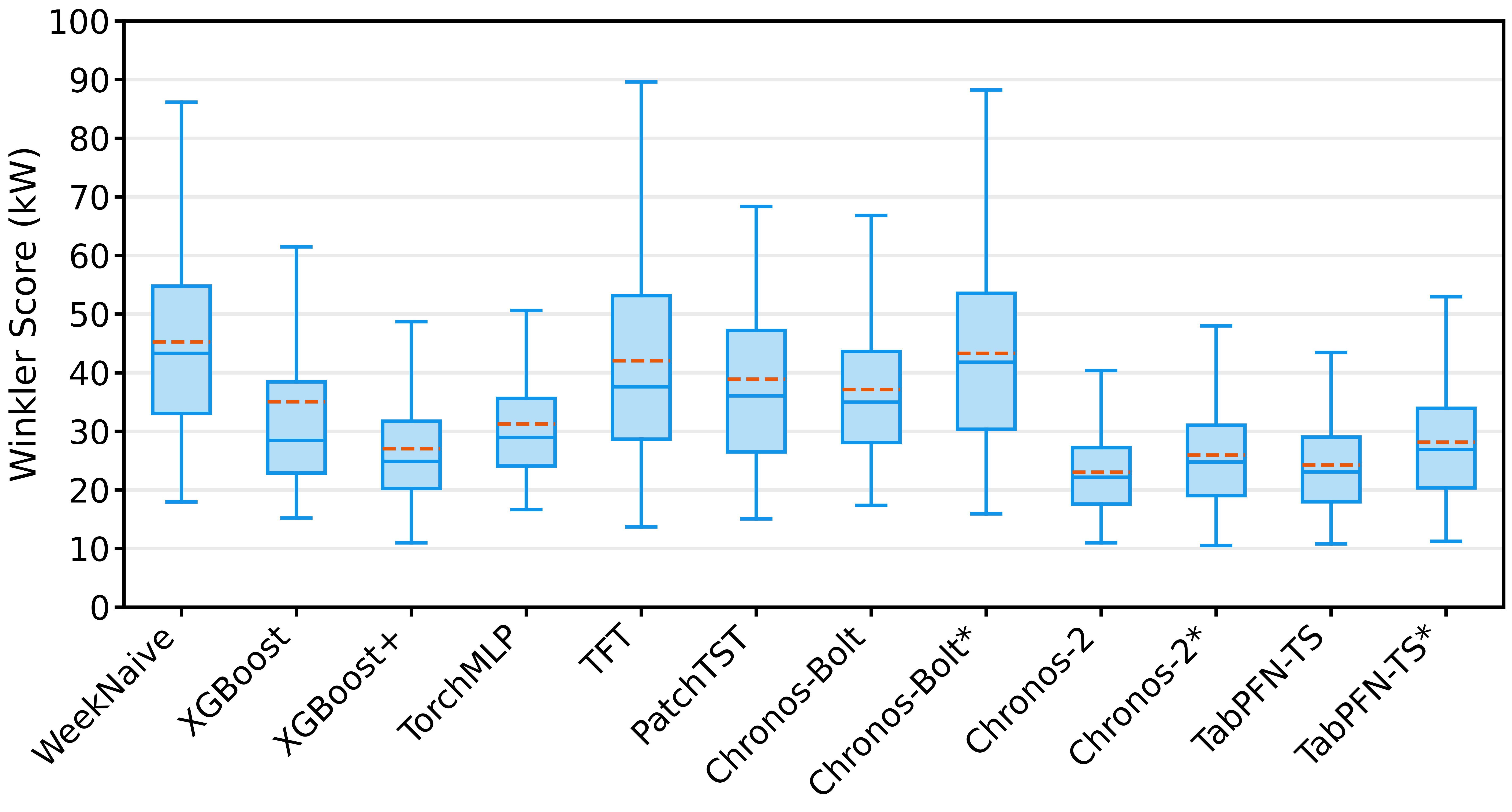}
    \caption{Winkler score distribution over feeders per model. The distribution is not over all timestamps or all forecast but over the metric per feeder and model. While Weeknaive and Chronos-Bolt perform worst at this metric judging by the median, their distribution is also the widest. The best-performing Chronos-2 and TabPFN-TS also exhibit more narrow metric spreads over the feeders.}
    \label{fig:wis_boxplot}
    \Description{Box plots of winkler score distributions over feeders for all models.}
\end{figure}



\end{document}
\endinput

%% file: acronyms.tex
\begin{acronym}[LV]\itemsep=-5pt
    \acro{LV}[LV]{low-voltage}
    \acro{MV}[MV]{medium-voltage}
    \acro{TSFM}[TSFM]{time-series foundation model}
    \acro{KPI}[KPI]{key performance indicator}
    \acro{DSO}[DSO]{distribution system operator}
    \acro{ML}[ML]{machine learning}
    \acro{PV}[PV]{photovoltaic}
    \acro{AI}[AI]{artificial intelligence}
    \acro{LLM}[LLM]{large language model}
    \acro{MLP}[MLP]{multilayer perceptron}
    \acro{DNN}[DNN]{deep neural network}
    \acro{MAE}[MAE]{mean absolute error}
    \acro{RMSE}[RMSE]{root mean square error}
    \acro{TFT}[TFT]{temporal fusion transformer}
    \acro{GAMLSS}[GAMLSS]{Generalised Additive Models for Location, Scale, and Shape}
\end{acronym}

%% file: tables/table_selection_timeseries_foundation_models.tex
\begin{table}[ht!]
\begin{threeparttable}
\centering
\small
\caption{Classification and selection of identified \acp{TSFM} based on various criteria. The availability of open weights, the support of probabilistic forecasting and covariates are prerequisites for \acp{TSFM} to be included in this study for \ac{LV} load forecasting. The models in \textbf{bold} are the final selection for this study.}
\Description{Table with time series foundation models categorized into different classes and the selected time series foundation models in this study Chronos-Bolt, Chronos-2 and TabPFN-TS in bold.}
\label{tab:selection_timeseries_foundation_models}

    \begin{tabular}{lccc|c}
        \toprule
        \textbf{Model} & \makecell[lt]{\textbf{Open}\\ \textbf{weights}} & \makecell[lt]{\textbf{Probabilistic}\\ \textbf{forecasting}} & \makecell[lt]{\textbf{Covariate}\\ \textbf{support}} & \makecell[lt]{\textbf{Fine-tuning}\\ \textbf{code}}\\
        \midrule
        Chronos         & \checkmark               & \checkmark      & Non-native\tnote{c}   & \checkmark   \\
        \textbf{Chronos-Bolt}    & \checkmark               & \checkmark      & Non-native\tnote{c}   & \checkmark  \\
        \textbf{Chronos-2}       & \checkmark               & \checkmark      & \checkmark                   & \checkmark\\
        Moirai          & \checkmark               & \checkmark       & (\checkmark)             & limited\\
        Moirai-MoE      & (\checkmark)\tnote{b}                & \checkmark      & (\checkmark)             & \\
        \textbf{TabPFN-TS}       & (\checkmark)\tnote{b}    & \checkmark      & \checkmark             & \checkmark\\
        \midrule
        FlowState       & \checkmark               & \checkmark       &                   & \\
        ForecastPFN     & \checkmark               &        &              & \\
        GPT4TS          & depends\tnote{a}  &       &              & \checkmark\\
        Kairos          & \checkmark               &        &              & \\
        Lag-Llama       & \checkmark               & \checkmark      &              & \checkmark\\
        LLM4TS          & depends\tnote{a}  &       &              & \checkmark\\
        Moirai-2        & (\checkmark)\tnote{b}                & \checkmark       &              & \\
        Moment          & \checkmark               &       &              & \checkmark\\
        RWKV-TS         & \checkmark               &       & \checkmark            &  \\
        Sundial         & \checkmark               & \checkmark       &              & \\
        TimeGPT         &                & \checkmark       & \checkmark             & \\
        Time-LLM        & depends\tnote{a}  &       &              & \checkmark\\
        Time-MoE        & \checkmark               &       &              & \checkmark\\
        Timer           & \checkmark               &       &              & \checkmark\\
        TimesFM         & \checkmark               &       & Non-native\tnote{c}      & \checkmark\\
        TTM             & \checkmark               &       & \checkmark             & \checkmark\\
        TiRex           & (\checkmark)\tnote{b}    & \checkmark       &              & \\
        Toto            & \checkmark               & \checkmark       &             & \\
        UniTS           & \checkmark               &       &              & \checkmark\\
        YingLong        & \checkmark               & \checkmark       &              & \\
        \bottomrule
    \end{tabular}

\begin{tablenotes}
\footnotesize
\item[a] Depends on base-LLM
\item[b] Non-permissive license
\item[c] Covariate regressor
\end{tablenotes}
\end{threeparttable}
\end{table}

%% file: tables/table_experimental_setup_models.tex
\begin{table*}[ht!]
\begin{threeparttable}
    \centering
    \caption{Experimental setup of the six baseline models (top) and three \acp{TSFM} (bottom) for load forecasting with four day forecast horizon with 15 minutes temporal resolution. Hence, one day of context (lagged features) corresponds to $96$ timestamps. Weather, time and metadata represent covariates, while \textit{H\&W} means that only holiday and weekday encodings are provided and no sine-cosine encoding because they are incorporated in the model. Zero-shot means that no model training is conducted and we use in-context learning instead. XGBoost in the second row is solely trained on metadata of the \ac{LV} feeder without lagged features of the time series. For hyperparameter configurations, see \Cref{tab:model_config_rest}. CPU refers to a 4-core 28 GB memory configuration. The GPU compute configuration is equipped with 8 cores, 56 GB memory, and an NVIDIA Tesla T4 GPU.}
    \Description{Table with the experimental setup for each model.}
    \label{tab:table_experimental_setup_models}
    \begin{tabular}{lrlrccccc}
        \toprule
        Model          & Ref. & Compute & Context & Scaling & Weather & Time & Metadata & Zero-shot \\ 
        \midrule
        WeekNaive      & & CPU     & 2688    &       &       &    &      & \checkmark         \\
        XGBoost        & \cite{chenXGBoostScalableTree2016, treutleinGeneratingPeakawarePseudomeasurements2025} & CPU     & -       & \checkmark     & \checkmark     & \checkmark  & \checkmark    &          \\
        XGBoost+ & \cite{chenXGBoostScalableTree2016} & CPU     & 384      & \checkmark     & \checkmark     & \checkmark  & \checkmark    &          \\
        TorchMLP       & \cite{Ansel_PyTorch_2_Faster_2024} & GPU     & 1344     & \checkmark     & \checkmark     & \checkmark  &      &          \\
        TFT            & \cite{limTemporalFusionTransformers2020} & GPU     & 1344     & \checkmark     & \checkmark     & \checkmark  & \checkmark    &          \\
        PatchTST       & \cite{nieTimeSeriesWorth2023} & GPU     & 1344     & \checkmark     & \checkmark     & \checkmark  &      &          \\ \midrule
        Chronos-Bolt   & \cite{ansariChronosLearningLanguage2024,abdulfatiransariFastAccurateZeroshot2024} & GPU     & 1344    &       & (\checkmark)\tnote{$\ast$}     & (\checkmark)\tnote{$\ast$} &      & \checkmark        \\
        Chronos-2      & \cite{ansariChronos2UnivariateUniversal2025} & GPU     & 1344    &       & (\checkmark)\tnote{$\ast$}     & (H\&W)\tnote{$\ast$} &      & \checkmark        \\
        TabPFN-TS      & \cite{hollmannTabPFNTransformerThat2023,hollmannAccuratePredictionsSmall2025,hooTablesTimeHow2025} & GPU     & 1344     &       & (\checkmark)\tnote{$\ast$}     & (H\&W)\tnote{$\ast$}  &      & \checkmark        \\ 
        \bottomrule
    \end{tabular}

\begin{tablenotes}
\footnotesize
\item[$\ast$] not for univariate configurations
\end{tablenotes}
\end{threeparttable}
\end{table*}

%% file: tables/table_metric_results_standard.tex
\begin{table*}[htb]
\begin{threeparttable}
\caption{Metric results for the six baseline models (top) and three \acp{TSFM} Chronos-Bolt, Chronos-2 and TabPFN-TS with and without covariates (bottom) considering point estimation (left) and quantile regression (right). Metric values are averaged for individual \ac{LV} feeders and then the median of the $200$ feeders is shown here. The considered quantiles are $0.05$ and $0.95$ (90 \% interval). The pinball loss incorporates the quantiles $0.05$ and $0.95$ and not the median. The best value per metric is in bold, second best is in italic. MAE, RMSE, pinball loss, Winkler score, and interval width are in kW.}
\label{tab:metric_results_standard}
\Description{Table with overall metric results of standard metrics for the six baselines and the three time-series foundation models with the variants with and without covariates.}
\centering
\begin{tabular}{@{}l|lll|lllll@{}}
\toprule
Model 
    & \parbox{1cm}{Median \newline MAE}
    & \parbox{1cm}{Median \newline RMSE}
    & \parbox{1cm}{Median \newline R²} 
    & \parbox{1cm}{Median \newline Pinball \newline Loss} 
    & \parbox{1cm}{Median \newline Winkler \newline Score}
    & \parbox{1cm}{Median \newline Interval \newline Width} 
    & \parbox{1cm}{Median \newline Empirical \newline Coverage} \\ 
\midrule
WeekNaive      & 5.315          & 7.384          & 0.1887            & 1.083    & 43.32       & 11.58 & 0.581                      \\
XGBoost & 5.02 & 6.81 & 0.3533 & 0.7107 & 28.43 & 17.11 & 0.8265 \\
XGBoost+ & 4.184 & 5.924 & 0.5116 & 0.6219 & 24.88 & 13.7 & 0.809 \\
TorchMLP & 4.673 & 6.45 & 0.4076 & 0.7235 & 28.94 & 22.76 & 0.9216 \\
TFT & 5.769 & 8.148 & 0.0793 & 0.9399 & 37.6 & 18.03 & 0.8 \\
PatchTST & 5.58 & 7.668 & 0.1939 & 0.901 & 36.04 & 17.56 & 0.8099 \\ \midrule
Chronos-Bolt & \textit{4.11} & \textit{5.726} & \textit{0.5349} & 0.8746 & 34.98 & \textbf{8.652} & 0.6211 \\
Chronos-Bolt\tnote{$\ast$} & 5.118          & 7.071          & 0.268  & 1.044   & 41.78    & \textit{10.8}           & 0.6388           \\ \midrule     
Chronos-2      & \textbf{3.839} & \textbf{5.468} & \textbf{0.569} & \textbf{0.5545} & \textbf{22.18} & 16.33          & \textbf{0.8975}  \\
Chronos-2\tnote{$\ast$} & 4.813          & 6.771          & 0.2789     & 0.6193       & 24.77    & 18.28          & 0.8844               \\ \midrule
TabPFN-TS      & 4.137 & 5.879          & 0.5085         & \textit{0.5767} & \textit{23.07}  & 18.86          & 0.9269           \\ 
TabPFN-TS\tnote{$\ast$}  & 4.996          & 7.127          & 0.2533   & 0.6726     & 26.9       & 20.36          & \textit{0.9105}           \\      
\bottomrule
\end{tabular}

\begin{tablenotes}
\footnotesize
\item[$\ast$] model inference without covariates (in particular without weather information)
\end{tablenotes}

\end{threeparttable}
\end{table*}

%% file: tables/table_metric_results_application_oriented.tex
\begin{table*}[htb]
\begin{threeparttable}
\caption{Application-oriented metric results for the six baseline models (top) and the three \acp{TSFM} with and without covariates (bottom) distinguished between consumers and producers. Recall, precision and F1 score are calculated for the model's point estimation and the appropriate quantile. The calculation is derived in \Cref{fig:metric_visualization_application_oriented}. The best value per metric is in bold, second best is in italics. Out of 34944 timestamps per feeder, 5470 are selected as consumer peaks and 691 as producer peaks from the ground truth on average. This means, on average, there are 15 consumer and 2 producer peaks per day in the ground truth. Note that these numbers vary between seasons and feeders.}
\label{tab:metric_results_application_oriented}
\Description{Table with metric results from confusion matrix of the application-oriented metric.}
\begin{tabular} {@{}lllllllllllll@{}}
\toprule
               & \multicolumn{6}{l}{Consumer}                                          & \multicolumn{6}{l}{Producer}                                          \\ \cmidrule(l){2-13} 
 &  \multicolumn{2}{l}{Precision} & \multicolumn{2}{l}{Recall} & \multicolumn{2}{l}{F1 Score} & \multicolumn{2}{l}{Precision} & \multicolumn{2}{l}{Recall} & \multicolumn{2}{l}{F1 Score} \\ \cmidrule(l){2-13}
Model & Point & 0.95 & Point & 0.95 & Point & 0.95 & Point & 0.05 & Point & 0.05 & Point & 0.05 \\ \midrule
WeekNaive & 0.7225 & 0.4854 & 0.6557 & 0.8904 & 0.6875 & 0.6283 & 0.554 & 0.3904 & 0.3277 & 0.8559 & 0.4118 & 0.5363 \\
XGBoost & 0.6962 & 0.3527 & 0.4968 & 0.9272 & 0.5798 & 0.511 & \textit{0.8124} & 0.4896 & 0.5297 & 0.9352 & 0.6412 & 0.6427 \\
XGBoost+ & 0.7647 & 0.4337 & 0.675 & 0.957 & 0.7171 & 0.5969 & 0.7876 & \textit{0.498} & 0.6092 & 0.9266 & 0.687 & \textit{0.6478} \\
TorchMLP & \textbf{0.8481} & 0.3379 & 0.4782 & 0.9739 & 0.6116 & 0.5018 & 0.8107 & 0.3222 & 0.4656 & 0.9204 & 0.5915 & 0.4773 \\
TFT & 0.6953 & 0.3544 & 0.5696 & 0.9437 & 0.6262 & 0.5153 & 0.5226 & 0.3625 & 0.2425 & 0.6156 & 0.3313 & 0.4563 \\
PatchTST & 0.7077 & 0.3911 & 0.6175 & 0.9391 & 0.6595 & 0.5522 & 0.5381 & 0.3768 & 0.2413 & 0.6969 & 0.3332 & 0.4891 \\ \midrule
Chronos-Bolt & 0.7846 & \textbf{0.5388} & 0.6768 & 0.9014 & 0.7267 & \textbf{0.6745} & \textbf{0.8363} & \textbf{0.6582} & \textit{0.648} & 0.8567 & \textit{0.7302} & \textbf{0.7445} \\
Chronos-Bolt* & 0.7341 & \textit{0.495} & 0.6664 & 0.9016 & 0.6986 & \textit{0.6391} & 0.5647 & 0.4095 & 0.455 & 0.8062 & 0.504 & 0.5431 \\ \midrule
Chronos-2 & \textit{0.7873} & 0.3836 & \textbf{0.7153} & \textit{0.9808} & \textbf{0.7496} & 0.5515 & 0.7942 & 0.4266 & \textbf{0.7738} & \textit{0.9897} & \textbf{0.7838} & 0.5962 \\
Chronos-2* & 0.7795 & 0.3841 & 0.674 & 0.9745 & 0.7229 & 0.551 & 0.5261 & 0.3181 & 0.6137 & 0.9802 & 0.5665 & 0.4803 \\ \midrule
TabPFN-TS & 0.775 & 0.3586 & \textit{0.6898} & \textbf{0.9816} & \textit{0.7299} & 0.5253 & 0.8118 & 0.3593 & 0.5999 & \textbf{0.9926} & 0.6899 & 0.5276 \\
TabPFN-TS* & 0.7415 & 0.3433 & 0.6741 & 0.9773 & 0.7062 & 0.5081 & 0.5585 & 0.3101 & 0.3886 & 0.9514 & 0.4583 & 0.4677  \\ \bottomrule
\end{tabular}

\begin{tablenotes}
\footnotesize
\item[$\ast$] model inference without covariates (in particular without weather information)
\end{tablenotes}

\end{threeparttable}
\end{table*}

%% file: tables/table_overview_timeseries_foundation_models.tex
\begin{table*}[ht!]
\begin{threeparttable}
    \centering
    \caption{Overview of identified \acp{TSFM} in alphabetical order. Information is derived from both the reference (Ref.) and the model's page on Hugging Face\tnote{a}. Academic and industrial refers to the organizations with which the authors are associated. A model may comprise variants with different parameter sizes (K $\mathrel{\hat{=}}$ kilo, M $\mathrel{\hat{=}}$ mega, B $\mathrel{\hat{=}}$ billion) which is indicated by specifying multiple parameters. The parameters of \acp{TSFM} requiring a base \ac{LLM} depend on the parameters of the \ac{LLM}~\cite{EpochAIModels2025}.}
    \label{tab:overview_timeseries_foundation_models}
    \Description{Table with an overview about all the identified time series foundation models.}
    \begin{tabular}{lllrrll}
        \toprule
        Model & Academic & Industrial & Year & Parameters & Architecture\tnote{b} & Ref. \\
        \midrule
        Chronos       & & \checkmark                              & 2024      & 8M - 710M               & encoder-decoder      & \cite{ansariChronosLearningLanguage2024}\\
        Chronos-Bolt  & & \checkmark                              & 2024      & 9M - 205M               & encoder-decoder      & \cite{abdulfatiransariFastAccurateZeroshot2024,ansariChronosLearningLanguage2024}\\
        Chronos-2     & & \checkmark                              & 2025      & 28M - 120M             & encoder-decoder      & \cite{ansariChronos2UnivariateUniversal2025}\\
        FlowState     & \checkmark & \checkmark                              & 2025      & 9M                    & state space model (SSM)      & \cite{grafFlowStateSamplingRate2025}\\
        ForecastPFN   & \checkmark & \checkmark                & 2023      & 1.5M                  & encoder              & \cite{dooleyForecastPFNSyntheticallyTrainedZeroShot2023}\\
        GPT4TS        & & \checkmark                             & 2023      & $\approx$ base LLM   & -                    & \cite{zhouOneFitsAll2023}\\
        Kairos        & \checkmark & \checkmark                 & 2025      & 10M - 50M         & encoder-decoder      & \cite{fengKairosAdaptiveGeneralizable2025}\\
        Lag-Llama     & \checkmark &                          & 2023      & 2.5M                  & decoder              & \cite{rasulLagLlamaFoundationModels2023}\\
        LLM4TS        & \checkmark &                          & 2024      & $\approx$ base LLM   & -                     & \cite{changLLM4TSAligningPreTrained2024}\\
        Moirai        & & \checkmark                          & 2024      & 14M - 311M        & encoder               & \cite{wooUnifiedTrainingUniversal2024}\\
        Moirai-2      & & \checkmark                          & 2025      & 11M - 305M        & decoder               & \cite{liuMoirai20When2025}\\
        Moirai-MoE    & & \checkmark                          & 2024      & 117M - 935M            & decoder               & \cite{liuMoiraiMoEEmpoweringTime2024}\\
        Moment        & \checkmark &                          & 2024      & 40M - 385M       & encoder               & \cite{goswamiMOMENTFamilyOpen2024}\\
        RWKV-TS       & \checkmark &                          & 2024      & 24M                   & RNN           & \cite{houRWKVTSTraditionalRecurrent2024}\\
        Sundial       & \checkmark &                          & 2025      & 128M                  & decoder                & \cite{liuSundialFamilyHighly2025}\\
        TabPFN-TS     & \checkmark & \checkmark                 & 2025      & 11M                   & encoder           & \cite{hollmannTabPFNTransformerThat2023,hollmannAccuratePredictionsSmall2025,hooTablesTimeHow2025}\\
        TimeGPT       & & \checkmark                              & 2023      & Not disclosed         & encoder-decoder        & \cite{garzaTimeGPT12024}\\
        TimeLLM       & \checkmark & \checkmark         & 2023      & $\approx$ base LLM    & -                      & \cite{jinTimeLLMTimeSeries2024}\\
        Time-MoE      & \checkmark & \checkmark               & 2024      & 113M - 2.4B    & decoder                & \cite{shiTimeMoEBillionScaleTime2025}\\
        Timer         & \checkmark &                          & 2024      & 85M                   & decoder                & \cite{liuTimerGenerativePretrained2024}\\
        TimesFM       & & \checkmark                              & 2024      & 200M - 500M            & decoder                & \cite{dasDecoderonlyFoundationModel2024}\\
        TTM           & & \checkmark                                 & 2024      & 800K                  & TSMixer                & \cite{ekambaramTinyTimeMixers2024}\\
        TiRex         & \checkmark & \checkmark                    & 2025      & 35M                   & xLSTM                  & \cite{auerTiRexZeroShotForecasting2025}\\
        Toto          & & \checkmark                             & 2025      & 151M                  & decoder               & \cite{cohenThisTimeDifferent2025}\\
        UniTS         & \checkmark &                          & 2024      & 1.5M - 8M   & encoder                & \cite{gaoUniTSUnifiedMultiTask2024}\\
        YingLong      & & \checkmark                             & 2025      & 6M - 300M               & encoder                & \cite{wangOutputScalingYingLongDelayed2025}\\
        \bottomrule
    \end{tabular}
    
    \begin{tablenotes}
    \footnotesize
    \item[a] https://huggingface.co
    \item[b] Whenever encoder, decoder or encoder-decoder are listed, they refer to transformer components.
    \end{tablenotes}
\end{threeparttable}
\end{table*}

%% file: tables/table_hyperparameter_models.tex
\begin{table}[htb]
\centering
\caption{Configuration and hyperparameters used for models in experiments as shown in \Cref{tab:table_experimental_setup_models}. All other model settings are default values.}
\label{tab:model_config_rest}
\begin{tabularx}{\linewidth}{@{}lX@{}}
\toprule
Model          & Configuration \\ 
\midrule
WeekNaive      & equal-weighted average of the last 4 weeks \\
XGBoost        & 1500 estimators, learning rate $3\mathrm{e}{-2}$ \\
XGBoost + Past & 1500 estimators, learning rate $3\mathrm{e}{-2}$, 12 last timestamps from the past are discarded \\
TorchMLP       & learning rate $5\mathrm{e}{-4}$; hidden layers: 4224 \& 2112; after each layer: activation function ReLU, dropout probability 0.1 \\
TFT            & learning rate $3.8\mathrm{e}{-4}$, hidden dimension 64, 1000 maximum epochs \\
PatchTST       & learning rate $3\mathrm{e}{-4}$, model dimension 32, 1000 maximum epochs  \\ \midrule
Chronos-Bolt   & "base", 200M parameters, CatBoost covariate regressor, target scaler "standard" \\
Chronos-2      & 120M parameters \\
TabPFN-TS      & 8 estimators, output selection "median", memory saving off, running index feature, calendar feature, auto-seasonal feature \\ 
\bottomrule
\end{tabularx}
\end{table}

%% file: tables/train_test_tikz.tex
\begin{tikzpicture}[
  x=0.5cm,   
  y=0.5cm,   
  >=Latex,
  every node/.style={align=center},
  thick
]

\def\xA{0}
\def\xB{5}
\def\xC{10}

\def\yBottom{0}   
\def\yMid{8}      
\def\yTop{10}     

\draw ( \xA, \yBottom ) rectangle ( \xC, \yTop );

\draw ( \xB, \yBottom ) -- ( \xB, \yTop ); 
\draw ( \xA, \yMid ) -- ( \xC, \yMid );    

\node at ($( \xA,\yBottom)!.5!( \xB,\yMid)$) {train};
\node at ($( \xB,\yBottom)!.5!( \xC,\yMid)$) {test (seen)};
\node at ($( \xA,\yMid)!.5!( \xB,\yTop)$) {ignored};
\node at ($( \xB,\yMid)!.5!( \xC,\yTop)$) {test (unseen)};

\draw[->] ( \xA, \yBottom ) -- ++(\xA, \yTop + 0.8) node[above] {Feeder ID};
\draw[->] ( \xA, \yBottom ) -- ++(\xC + 1.5, 0) node[right] {Time};

\node[left] at ( \xA, \yBottom + 0.2) {1};
\node[left]       at ( \xA, \yMid )    {161};
\node[left]       at ( \xA, \yTop )    {200};

\node[below] at (\xA, \yBottom) {2023/04/01};
\node[below] at (\xB, \yBottom) {2024/04/01};
\node[below] at (\xC, \yBottom) {2025/04/01};

\end{tikzpicture}

%% file: tables/table_metric_results_peakmetrics.tex
\begin{table*}[htb]
\begin{threeparttable}
\caption{Peak metric results for the six baseline models (top) and the three \acp{TSFM} with and without covariates (bottom). Peak detection is based on the peak metric variant proposed in \Cref{subsec:application_oriented_metric} (\Cref{subfig:metric_visualization_application_oriented_1} and \Cref{subfig:metric_visualization_application_oriented_2}). Metric values are averaged for individual \ac{LV} feeders and then the median of the $200$ feeders is shown here. The considered quantiles are $0.05$ and $0.95$ (90~\% interval). The pinball loss incorporates the quantiles $0.05$ and $0.95$ and not the median. The best value per metric is in bold, second best is in italic. MAE, pinball loss, Winkler score, and interval width are in kW.}
\label{tab:metric_results_peakmetric}
\Description{Table with metric results of the standard metrics evaluated only for identified peaks.}
\begin{tabular} {@{}lllllllllll@{}}
\toprule
 &
  \multicolumn{2}{p{1.8cm}}{\parbox{1.8cm}{Median \newline MAE}} &
  \multicolumn{2}{p{1.8cm}}{\parbox{1.8cm}{Median \newline Pinball \newline Loss}} &
  \multicolumn{2}{p{1.8cm}}{\parbox{1.8cm}{Median \newline Winkler \newline Score}} &
  \multicolumn{2}{p{1.8cm}}{\parbox{1.8cm}{Median \newline Interval\newline Width}} &
  \multicolumn{2}{p{1.8cm}}{\parbox{1.8cm}{Median \newline Empirical\newline Coverage}} \\ \cmidrule(l){2-11}
Model          & Cons.           & Prod. & Cons.           & Prod.            & Cons.           & Prod.           & Cons.            & Prod.            & Cons.           & Prod.           \\ \midrule
WeekNaive & 8.375 & 18.65 & 2.028 & 3.402 & 81.11 & 136.1 & 13.72 & 34.47 & 0.4929 & 0.5161 \\
XGBoost & 9.739 & 12.8 & 1.351 & 1.709 & 54.04 & 68.38 & 20.19 & 31.44 & 0.6572 & 0.6479 \\
XGBoost+ & 7.931 & 12.37 & 1.384 & 2.069 & 55.37 & 82.77 & 17.55 & 30.05 & 0.6721 & 0.6256 \\
TorchMLP & 9.578 & 15.58 & 1.345 & 2.33 & 53.79 & 93.2 & 26.18 & 48.28 & 0.7487 & 0.6444 \\
TFT & 9.642 & 24.79 & 1.534 & 6.284 & 61.36 & 251.3 & 21.32 & 35.13 & 0.6851 & 0.3538 \\
PatchTST & 8.883 & 22.26 & 1.664 & 4.915 & 66.57 & 196.6 & 19.85 & 37.36 & 0.6218 & 0.4582 \\ \midrule
Chronos-Bolt & 7.999 & \textit{11.74} & 2.105 & 3.251 & 84.19 & 130 & \textbf{10.41} & \textbf{15.47} & 0.4276 & 0.4092 \\
Chronos-Bolt\tnote{$\ast$} & 8.534 & 19.14 & 2.145 & 4.46 & 85.81 & 178.4 & \textit{12.19} & \textit{28.15} & 0.4454 & 0.4768 \\ \midrule
Chronos-2 & \textbf{7.304} & \textbf{8.786} & \textit{1.091} & \textbf{1.278} & \textit{43.64} & \textbf{51.12} & 20.28 & 33.76 & 0.7865 & \textit{0.8595} \\
Chronos-2\tnote{$\ast$} & 8.063 & 14.21 & 1.17 & 1.893 & 46.81 & 75.73 & 21.71 & 47.95 & 0.7684 & 0.7934 \\ \midrule
TabPFN-TS & \textit{7.431} & 13.07 & \textbf{1.042} & \textit{1.475} & \textbf{41.69} & \textit{58.99} & 22.71 & 44.72 & \textbf{0.81} & \textbf{0.889} \\
TabPFN-TS\tnote{$\ast$} & 8.184 & 20.41 & 1.135 & 2.333 & 45.39 & 93.32 & 23.71 & 48.48 & \textit{0.8043} & 0.7762   \\ \bottomrule
\end{tabular}

\begin{tablenotes}
\footnotesize
\item[$\ast$] model inference without covariates (in particular without weather information)
\end{tablenotes}

\end{threeparttable}
\end{table*}

%% file: tables/table_runtimes.tex
\begin{table}[htb]
\begin{threeparttable}
    \caption{Training and inference times of the baseline models (top) and \acp{TSFM} with and without covariates (bottom). Total times are in the format hh:mm:ss, and per-forecast times are in ms. See \Cref{tab:table_experimental_setup_models} for the compute configurations.}
    \label{tab:runtimes}
    \Description{Table with the runtimes for training and inference for each model.}
    \centering
    \begin{tabular} {@{}lllll@{}}
        \toprule
                  & \multicolumn{2}{p{2.5cm}}{Training Time}     & \multicolumn{2}{p{2.5cm}}{Inference Time} \\  \cmidrule(l){2-5}
        Model      & Total & Per Forecast & Total & Per Forecast \\ \midrule
        WeekNaive      & \textbf{00:00:00} & \textbf{0} & \textbf{00:00:56} & \textbf{3.1}         \\
        XGBoost        & 00:37:29 & 123.6       & 00:14:13 & 46.9                              \\
        XGBoost+ & 04:43:14 & 933.7       & 02:03:32 & 407.3                          \\
        TorchMLP       & 00:59:27 & 196.0        & \textit{00:01:52} & \textit{6.2}                       \\
        TFT            & 00:15:16 & 50.3      & 00:03:25 & 11.3                             \\
        PatchTST       & 00:09:56 & 32.7         & 00:03:48 & 12.5                     \\ \midrule
        Chronos-Bolt   & 00:09:00 & 29.7         & 01:25:50 & 283.0                            \\
        Chronos-Bolt\tnote{$\ast$} & \textbf{00:00:00} & \textbf{0}         & 01:28:14 & 290.9                            \\ \midrule
        Chronos-2      & \textbf{00:00:00} & \textbf{0} & 00:24:45 & 81.6                \\
        Chronos-2\tnote{$\ast$} & \textbf{00:00:00} & \textbf{0}   & 00:03:20 & 11.0                            \\ \midrule
        TabPFN-TS      & \textbf{00:00:00} & \textbf{0} & 08:07:44 & 1607.9                \\ 
        TabPFN-TS\tnote{$\ast$} & \textbf{00:00:00} & \textbf{0}    & 07:13:26 & 1428.9                            \\ \bottomrule 
    \end{tabular}

\begin{tablenotes}
\footnotesize
\item[$\ast$] model inference without covariates (in particular without weather information)
\end{tablenotes}
\end{threeparttable}
\end{table}